\documentclass{article}

\usepackage{PRIMEarxiv}

\usepackage[utf8]{inputenc} % allow utf-8 input
\usepackage[T1]{fontenc}    % use 8-bit T1 fonts
\usepackage{hyperref}       % hyperlinks
\usepackage{url}            % simple URL typesetting
\usepackage{booktabs}       % professional-quality tables
\usepackage{amsfonts}       % blackboard math symbols
\usepackage{nicefrac}       % compact symbols for 1/2, etc.
\usepackage{microtype}      % microtypography
\usepackage{lipsum}
\usepackage{fancyhdr}       % header
\usepackage{algorithm}
\usepackage{algpseudocode}
\usepackage[pdftex]{graphicx}
\graphicspath{{figures/}}
\usepackage{caption}
\usepackage{wrapfig}
\usepackage{amsmath}
\usepackage{amsfonts}
\usepackage{enumitem}
\usepackage{textgreek}
\usepackage{xcolor}
\usepackage{float}
\usepackage[style=numeric]{biblatex}

\newcommand{\fedavg}{\texttt{FedAvg}}
\newcommand{\fedadam}{\texttt{FedAdam}}
\newcommand{\qfedavg}{\texttt{q-FedAvg}}
\newcommand{\ours}{\texttt{AdaFedAdam}}
\newcommand{\adam}{\texttt{Adam}}
\newcommand{\sgd}{\texttt{SGD}}
\newcommand{\adagrad}{\texttt{AdaGrad}}
\newcommand{\fedopt}{\texttt{FedOpt}}
\newcommand{\naccadam}{\texttt{\(N\)-AccAdam}}
\newcommand{\fednova}{\texttt{FedNova}}

\newcommand{\header}[1]{\textbf{\textsc{#1}}}

\newtheorem{theorem}{\textbf{Theorem}}

\newcommand{\change}[1]{{#1}}

\usepackage{soul}
\newcommand{\remove}[1]{}

\title{Accelerating Fair Federated Learning:\\Adaptive Federated Adam
}

\author{
  Li Ju, Tianru Zhang, Salman Toor, Andreas Hellander \\
  Department of Information Technology \\
  Uppsala University \\
  Uppsala, Sweden\\
  \texttt{\{li.ju, tianru.zhang, salman.toor, andreas.hellander\}@it.uu.se} \\
}

\begin{document}
\maketitle

\begin{abstract}
Federated learning is a distributed and privacy-preserving approach to train a statistical model collaboratively from decentralized data of different parties. However, when datasets of participants are not independent and identically distributed (\emph{non-IID}), models trained by naive federated algorithms may be biased towards certain participants, and model performance across participants is non-uniform. This is known as the fairness problem in federated learning. In this paper, we formulate fairness-controlled federated learning as a dynamical multi-objective optimization problem to ensure fair performance across all participants. To solve the problem efficiently, we study the convergence and bias of \adam{} as the server optimizer in federated learning, and propose Adaptive Federated Adam (\ours{}) to accelerate fair federated learning with alleviated bias. We validated the effectiveness, Pareto optimality and robustness of \ours{} in numerical experiments and show that \ours{} outperforms existing algorithms, providing better convergence and fairness properties of the federated scheme. 
\end{abstract}

% keywords can be removed
% \keywords{Federated Learning \and Fairness Problem \and Federated Optimization}

\section{Introduction}
% general introduction for FL
Federated Learning (FL), first proposed by \cite{mcmahan2017communication}, is an emerging collaborative learning technique enabling multiple parties to train a joint machine learning model with input privacy being preserved. By iteratively aggregating local model updates done by participating clients using their local, private data, a joint global model is obtained. The promise of federated learning is that this global model will have superior performance compared to the models that could be obtained by each participant in isolation. Compared with traditional distributed machine learning, FL works with larger local updates and seeks to minimize communication cost while keeping the data of participants local and private. With increasing concerns about data security and privacy protection, federated learning has attracted much research interest (\cite{kairouz2021advances, li2020federated}) and has been proven to work effectively in various application domains (\cite{li2020review, xu2021federated}).

% talk about model fairness in this paragraph
When the datasets at the client sites are not independent and identically distributed (IID), the standard algorithm for federated learning, \fedavg{}, can struggle to achieve  good model performance, with an increase of communication rounds (\cite{zhu2021federated, li2019convergence}) needed for convergence. Moreover, the global model trained with heterogeneous data can be biased towards some of the participants, while performing poorly for others (\cite{mohri2019agnostic}). This is known as unfairness problem in federated learning. There are ways to improve fairness in federated learning, at the cost of model convergence (\cite{mohri2019agnostic, li2019fair, hu2022federated, li2021ditto}). This study aims to contribute to the understanding of how to enable fair federated learning without negatively impacting the convergence rate. 
% It is still under-explored how to eliminate the unfairness without slowing down the convergence rate. 

Acceleration techniques for federated learning aim at reducing the communication cost and improving convergence. For instance, momentum-based and adaptive optimization methods such as \adagrad{}, \adam{}, \texttt{Momentum} \sgd{}) have been applied to accelerate the training process \cite{karimireddy2020mime, wang2019slowmo, reddi2020adaptive}). However, default hyperparameters of adaptive optimizers tuned for centralized training do not tend to perform well in federated settings (\cite{reddi2020adaptive}). Furthermore, optimal hyperparameters are not generalizable for federated learning, and hyperparameter optimization with e.g. grid search are needed for each specific federated task, which is infeasible due to the expensive (and sometimes unbounded) nature for federated learning. Further research is need to understand how to adapt optimizers for federated learning with minimal hyperparameter selection. 

In this study, to accelerate the training of fair federated learning, we 
% \change{analyze the convergence, bias and unfairness of \adam{} in federated settings and tackle the reasons why \fedadam{} fails to work as expected without fine-tuning first. Based on the analysis, we propose \ours{}, an adaptive algorithm for fair federated optimization with better convergence and fairness compared to \fedavg{}-based algorithms. }
formulate fairness-aware federated learning as a \emph{dynamical multi-objective optimization problem (DMOO)} problem.
By analyzing the convergence and bias of federated \adam{}, we propose Adaptive Federated Adam (\ours{}) to solve the formulated DMOO problem efficiently.
With experiments on standard benchmark datasets, we illustrate that \ours{} alleviates model unfairness and accelerates federated training. In additional, \ours{} is proved to be robust against different levels of data and resource heterogeneity, which suggests that the its performance on fair federated learning can be expected in real-life use cases. 

The remainder of the paper is structured as follows. Section \ref{sec:related_work} summarizes related work including different acceleration techniques for federated training and the fairness problem in federated learning. Then fair federated learning problem is formulated in Section \ref{sec:background} and Federated \adam{} is analyzed in Section \ref{sec:ana_fedadam}. Section \ref{sec:adafedadam} introduces the design of \ours{}. Setups and results of empirical experiments are presented in Section \ref{sec:experiments}. Finally, section \ref{sec:conclusion} concludes the paper and suggests future research directions.

\section{RELATED WORK}\label{sec:related_work}
In this section, we review recent techniques to accelerate federated training as well as studies of model fairness in FL. 

\subsection{Acceleration techniques for federated learning}
Adaptive methods accelerate centralized training of neural networks over vanilla \sgd{} (\cite{kingma2014adam, zeiler2012adadelta}). In the context of federated learning, \cite{hsu2019measuring} and \cite{wang2019slowmo} introduced first-order momentum to update the global model by treating local updates as pseudo-gradients, showing the effectiveness of adaptive methods for federated learning. Further, \cite{reddi2020adaptive} demonstrated a two-level optimization framework \fedopt{} for federated optimization. On the local level, clients optimize the local objective functions while local updates are aggregated as "pseudo-gradients" to update the global model on the server level. From the view of the \fedopt{} framework, \fedavg{} uses \sgd{} as its local solver and \texttt{Gradient Descent} with a learning rate of 1 as the server optimizer. By applying adaptive optimizers (e.g. \adam{}) as the server optimizer, we obtain adaptive federated optimizers (e.g. \fedadam{}). It has been empirically validated that adaptive federated optimizers are able to accelerate training, however they need careful fine-tuning (\cite{reddi2020adaptive}). 

Fine-tuning for server optimizers is challenging for the following reasons: 
\begin{itemize}
    \item Due to the inherent differences of federated and centralized training, default hyperparameters of optimizers which work well on centralized training does not necessarily have satisfactory performances in federated training. 
    \item For adaptive optimizers, grid search needs to be done to get multiple hyperparameters optimized (\cite{reddi2020adaptive}), which is prohibitively expensive considering the orchestration cost for the entire federation. 
    \item The optimal hyperparameters for server-side optimizers are not generalizable between different federated tasks, and fine-tuning must be done for each individual task. 
\end{itemize}
It would greatly ease the use of server-side optimizers if the selection of hyperparameters were automatic. The proposed methods \ours{} minimizes the efforts of fine-tuning by adapting default hyperparameters of \adam{} in centralized settings to federated training. 

\subsection{Model fairness}
The concept of model unfairness describes the differences of model performance across participants in a federated training process. It was firstly highlighted by \cite{mohri2019agnostic}. Most federated learning algorithms optimize the global model solely focusing on the averaged loss. However, model performances between clients are not uniform when data between participants are heterogeneous, and the global model can be biased towards some participants. To reduce the unfairness, \cite{mohri2019agnostic} proposed the algorithm \texttt{Agnostic Federated Learning}, a minimax optimization approach that only optimizes the single device with the worst performance. Inspired by fair resource allocation, \cite{li2019fair} formulated fair federated learning as a fairness-controlled optimization problem with \(\alpha\)-fairness function (\cite{mo2000fair}). By increasing \(\alpha\) (\(q\) in the article), the desired fairness between participants can be improved at a cost of convergence. The algorithm \qfedavg{} was proposed to solve the optimization problem, which dynamically adjust step sizes of local \sgd{} by iteratively estimating Lipschitz constants. More recently, \cite{hu2022federated} interpreted federated learning as a multi-objective optimization problem, and adapted Multi-Gradient Descent Algorithm (MGDA) to federated settings as \texttt{FedMGDA+} to reduce the unfairness. Alternatively, \cite{li2021ditto} proposed \texttt{Ditto} to improve the performance fairness by personalizing global models on client sites. 

Unlike previous work that are based on \fedavg{} with improved fairness at a cost of model convergence, the here proposed approach \change{formulates fair federated learning as a dynamic multi-objective function and proposes \ours{} to solve the formulated problem. Compared with other \fedavg{}-based algorithms for fairness control, \ours{} offer equivalent fairness guarantee with  improved convergence properties. }
% --------------

\section{PRELIMINARIES \& PROBLEM FORMULATION}\label{sec:background}
\subsection{Standard federated learning}
Considering the distributed optimization problem to minimize the global loss function \(F(\mathbf{x})\) across \(K\) clients as follows: 
\begin{equation}\label{eq:fl}
    \min_{\mathbf{x}}[F(\mathbf{x}) := \sum_{k=1}^{K} p_{k} F_{k}(\mathbf{x})]
\end{equation}
where \(\mathbf{x}\) denotes the parameter set of function \(F\), \(F_k(\mathbf{x})\) is the local objective function of client \(k\) w.r.t local dataset \(D_k\), and \(p_k := \frac{|D_k|}{\sum |D|}\) denotes the relative sample size of \(D_k\) with number of samples \(|D_k|\) in \(D_k\). The data distribution on client \(k\) is denoted by \(\mathcal{D}_k\). 

\remove{
One of the baseline algorithm to solve the federated optimization, \fedavg{}, is noted as follows: 
\begin{equation}
    \mathbf{x}^{t+1} = \sum_{k=1}^{K^t} p_{k} \mathbf{x}^{t}_{k} \text{, where } \mathbf{x}^{t}_{k} := \textbf{SGD}(\mathbf{x}^t, \nabla_{\zeta \sim \mathcal{D}_k}, \eta, \mathcal{S}_k)
\end{equation}
In communication round \(t\), a subset of clients \(K^t \subseteq K\) participates the training. For all \(k \in [K^t]\), client \(k\) runs (mini-batch) \sgd{} with learning rate \(\eta\) on the global model \(\mathbf{x}^t\) for \(\mathcal{S}_k\) steps. Gradient estimation of each step \(\nabla_{\zeta \sim \mathcal{D}_k}\) is based on batches \(\zeta\) sampled from the local distribution \(\mathcal{D}_k\). Parameters of all local models are averaged weighted by the relative size of their local datasets \(p_k\) to get the global model \(\mathbf{x}^{t+1}\). 

Federated learning algorithms enable an two-level optimization structure with a client- and server-level\cite{reddi2020adaptive}. At the client level, local solvers are not limited to SGD to minimize the local loss functions. At the server level, aggregated local updates can be regarded as "pseudo-gradients" of the global model, and can be the input for the server solver to update the global model. This type of two-level optimization includes \fedadam{} and \texttt{FedAdaGrad}, which is obtained by applying \adam{} (\cite{kingma2014adam}) or \adagrad{} (\cite{lydia2019adagrad}) as the server optimizer. 
}
\subsection{Fair federated learning}
\change{
When \(\mathcal{D}_k\) are not identical across clients (the non-IID case), the standard formulation of federated learning can suffer from a significant fairness problem (\cite{mohri2019agnostic}) in addition to a potential loss of convergence. To improve  fairness, federated learning can be formulated with an \(\alpha\)-fairness function as \(\alpha\)-Fair Federated Learning (also known as \(q\)-Fair Federated Learning in \cite{li2019fair}) as follows: }

\begin{equation}\label{eq:ffl}
    \min_{\mathbf{x}}[F(\mathbf{x}) := \sum_{k=1}^{K} \frac{p_{k}}{\alpha+1} F^{\alpha+1}_{k}(\mathbf{x})]
\end{equation}

\change{
where notations is as in \eqref{eq:fl}. With the additional hyperparameter \(\alpha\), \(\alpha\)-Fair Federated Learning is able to control the desired fairness level in federated learning. Setting larger values for \(\alpha\) indicates greater demand for fair/uniform performances across clients at a cost of convergence (\cite{mo2000fair}), and setting \(\alpha=0\) reduces \(\alpha\)-Fair Federated Learning to the standard formulation of Federated Learning in \eqref{eq:fl}. 
}

\change{
It is challenging to solve the problem with distributed first-order optimization. With only access to gradients of local objective functions \(\nabla^t F_k(\mathbf{x})\), the gradient of \(F(\mathbf{x})\) at \(\mathbf{x}^t\) and the update rule of distributed \sgd{} are given as follows:}
\begin{align}\label{eq:grad_ffl}
    \nabla F(\mathbf{x}^t) &= \sum_{k=1}^{K} p_{k} F^{\alpha}_{k}(\mathbf{x}^t) \nabla F_k(\mathbf{x}^t) \\
    \mathbf{x}^{t+1} &:= \mathbf{x}^t - \eta \nabla F(\mathbf{x}^t)
\end{align}

\change{
It is noticeable that the gradient \(\nabla F(\mathbf{x}^t)\) has decreasing scales due to the factor \(F^\alpha(\mathbf{x}^t)\). As the number of iterations \(t\) increases, a decreasing \(F^\alpha(\mathbf{x}^t)\) scales gradients \(\nabla F(\mathbf{x}^t)\) down drastically. With a fixed learning rate \(\eta\), the update of \sgd{} \(-\eta \nabla F(\mathbf{x}^t)\) scales down correspondingly and thus, the convergence deteriorates. To improve the convergence, \cite{li2019fair} proposes \qfedavg{} to adjust learning rates adaptively. However, the convergence of \qfedavg{} is not satisfying since 1). the intrinsic challenge of optimizing the learning rate adaptively in \(\alpha\)-Fair Federated Learning still exists and 2). \fedavg{}-based \qfedavg{} does not utilize acceleration techniques. 
}

\subsection{Problem formulation}
\remove{
In real life use cases, participants of federated learning can be expected to join or leave the training at any time according to their availability in either cross-silo or cross-device scenarios. It is important to not only obtain a fair model after the training completes, but also keep the global model fair during the training process. The latter guarantees local performances whenever participants exit the training. Fairness during the training also encourages clients to contribute to the training continuously. 
}
\change{
In the field of multi-task learning, neural networks are designed to achieve multiple tasks at the same time by summing multiple component objective functions up as a joint loss function. 
}
In a similar spirit to fair federated learning, training multitask deep neural networks also requires to keep similar progress for all component objectives. Inspired by \cite{chen2018gradnorm}, we formulate fair federated learning as a dynamic multi-objective optimization problem (DMOP) in the following form: 

\begin{equation}\label{eq:iffl}
    \min_{\mathbf{x}}[F(\mathbf{x}, t) := 
    \frac{\sum_{k=1}^{K} p_{k} I^\alpha_k(t) F_{k}(\mathbf{x})}
    {\sum_{k=1}^{K} p_{k} I^\alpha_k(t)} ]
\end{equation}

where \(p_k\) is the size of dataset on client \(k\), and \(F_k(\mathbf{x})\) is the local objective function of client \(k\). Additionally, \textit{inverse training rate} is defined as \(I_k(t) := F_k(\mathbf{x}^t)/F_k(\mathbf{x}^0)\) for participant \(k\) at round \(t\), to quantify its training progress. \(\alpha \geq 0\) is a hyperparameter to adjust the model fairness similar to \(\alpha\) in \(\alpha\)-fairness function. % \(I^\alpha_k(t)\) is a dynamic weight for local objective function \(F_k(\mathbf{x})\) at round \(t\). 
The problem reduces to the federated optimization without fairness control if setting \(\alpha=0\), and it restores the minimax approach for multi-objective optimization (\cite{mohri2019agnostic}) if setting \(\alpha\) a sufficiently large value. 

\change{
Compared with \(\alpha\)-Fair Federated Learning, the proposed formulation has equivalent fairness guarantee without the problem of decreasing scales of gradients. Considering that the global model \(\mathbf{x}^0\) is initialized with random weights, we assume that \(F_{i}(\mathbf{x}^0) = F_{j}(\mathbf{x}^0)\) for \(\forall i, j \in [K]\). Then we have that the gradient of \(F(\mathbf{x}, t)\) at \(\mathbf{x}^t\) is given by: 
}
\begin{align}\label{eq:grad_iffl}
    \nabla F(\mathbf{x}^t, t) = \frac{\sum_{k=1}^{K} p_{k} F^{\alpha}_{k}(\mathbf{x}^t) \nabla F_k(\mathbf{x}^t)}{\sum_{k=1}^{K} p_{k} F^{\alpha}_{k}(\mathbf{x}^t)} \propto \sum_{k=1}^{K} p_{k} F^{\alpha}_{k}(\mathbf{x}^t) \nabla F_k(\mathbf{x}^t)
\end{align}

\change{
The gradient \(F(\mathbf{x}^t, t)\) of the DMOP formulation is proportional to the gradient of the \(\alpha\)-Fair Federated Learning (\eqref{eq:ffl}). Thus, with first-order optimization methods, the solution of the DMOP formulation is also the solution of the \(\alpha\)-fairness function, which has been proved to enjoy \((p, \alpha)\)\textit{-Proportional Fairness} (\cite{mo2000fair}). Moreover, the DMOP formulation of Fair Federated Learning does not have the problem of decreasing gradient scales in the \(\alpha\)-fairness function, so that distributed first-order optimization methods can be applied to solve the problem more efficiently. 
}

\section{ANALYSIS OF \textit{FEDADAM}}\label{sec:ana_fedadam}
In this section, we analyze the performance of \adam{} as the server optimizer in federated learning. We first study the effect of using accumulated updates as pseudo-gradients for \adam{} in centralized training. The bias introduced by averaging accumulated local updates without normalization in \fedadam{} is then discussed. 

\subsection{From \adam{} to \fedadam{}}
As the de facto optimizer for centralized deep learning, \adam{} provides stable performance with little need of fine-tuning. The pseudo code of \adam{} is shown in \ref{sec:pseudocode} 3. \adam{} provides adaptive stepsize selection based on the initial stepsize \(\eta\) for each individual coordinate of model weights. The adaptivity of stepsizes can be understood as continuously establishing \emph{trust regions} based on estimations of the first- and second-order momentum (\cite{kingma2014adam}), which are updated by exponential moving averages of gradient estimations and their squares with hyperparameters \(\beta_1\) and \(\beta_2\) in each step. 

The choice of hyperparameters in \adam{} can be explained by the certainty of directions for model updates. In centralized \adam{}, directions for updates are from gradient estimations \(\nabla_{\zeta \sim \mathcal{D}} F(\mathbf{x})\) obtained from a small batch of data \(\zeta\) with large variances, indicating low certainty of update directions. Thus, large \(\beta_1\) and \(\beta_2\) (\(0.9\) and \(0.999\) by default) are set to assign less weight for each gradient estimation when updating first- and second-order momentum. Low certainty of update directions also only allow small \emph{trust regions} to be constructed from small initial stepsize \(\eta\) (\(0.001\) by default). 

In federated learning, \fedadam{} is obtained if we apply \adam{} as the server optimizer and the size-weighted average of clients' local updates at round \(t\), \(\Delta_t\), acts as the pseudo-gradient. Although empirical results have shown that \fedadam{} outperforms the standard \fedavg{} with careful fine-tuning in terms of average loss (\cite{reddi2020adaptive}), several problems exist in \fedadam{}. In the following subsections, we analyze the problem of convergence loss of \fedadam{} and bias of pseudo-gradients used for \fedadam{}. 

\subsection{Adam with accumulated updates}
When data between clients are statistically homogeneous, the average of local updates is an unbiased estimator of accumulated updates of multiple centralized \sgd{} steps. Therefore, in IID cases, \fedadam{} shrinks as \adam{} with gradient estimation given by accumulated updates of \(N\) \sgd{} steps (\naccadam{}). Pseudo-codes of \naccadam{} is given in \ref{sec:pseudocode}. We prove that even in centralized settings, \naccadam{} has less convergence guarantee than standard \adam{} with same hyperparameters.

\begin{theorem}[Convergence of \naccadam{}]\label{thrm:accadam}
% \small
\textit{Assume the \(L\)-smooth convex loss function \(f(\mathbf{x})\) has bounded gradients \(\|\nabla\|_{\infty} \leq G\) for all \(\mathbf{x} \in \mathbb{R}^d\). Hyperparameters \(\epsilon\), \(\beta_2\) and \(\eta\) in \naccadam{} are chosen with the following conditions: \(\eta \leq \epsilon/2L\) and \(1-\beta_2 \leq \epsilon^2/16G^2\). The accumulated SGD updates at step \(t\) \(\Delta\mathbf{x}^t := -\sum_{n=1}^{N} \Delta_{n}\mathbf{x}^t /\eta_s\) is applied to Adam, where \(\Delta_n \mathbf{x}^t\) denotes the local update after \(n\) SGD steps with a learning rate of \(\eta_s\) on model \(\mathbf{x}^t\). \sgd{} exhibits approximately linear convergence with constants \((A, c)\). In the worst case, the algorithm has no convergence guarantee. In the best cases where \(R_t = N\) for all \(t\in [T]\), the converge rate is given by: }

\begin{equation}\label{eq:accadam}
 \frac{1}{T} \sum_{t=1}^T \|\nabla f(\mathbf{x}^t)\|^2 \leq \frac{f(\mathbf{x}^1) - f(\mathbf{x}^*)(\sqrt{\beta_2}G+\epsilon)}{\underbrace{(\frac{Nc}{1-(1-c)^N} - \frac{1}{2})}_{S}\eta T} 
\end{equation}
where \(R_t := \min | \frac{\Delta_{t,i}}{\nabla_{t,i}} |\) for \(i \in [d]\)
\end{theorem}

The proof for Theorem \ref{thrm:accadam} is deferred to \ref{subsec:proof1}. In the best case where \(R_t = N\) (which is almost not feasible), \naccadam{} gains \(S\) speedup compared with \adam{}. However, the computation cost of \naccadam{} is linear to \(N\) but the speedup \(S\) is sublinear to \(N\). Thus, with a fixed computation budget, the convergence rate of \naccadam{} is slower than \adam{} with the same hyperparameters. Compared with gradient estimation by a small batch of data, accumulated updates of multiple \sgd{} steps have larger certainty about directions of updates for the global model. To improve the convergence of \naccadam{}, it is possible to construct larger \emph{trust regions} with larger stepsize \(\eta\) and smaller \(\beta\)s with accumulated updates. 

\subsection{Bias of pseudo-gradients for \fedadam{}}
In federated settings, when data among clients are heterogeneous, averaging all local updates weighted by sizes of client datasets introduces bias toward a portion of clients. The biased pseudo-gradients lead to even lower convergence and increase the unfairness of \fedadam{}. 

\cite{wang2020tackling} has proved that there exists objective inconsistency between the stationary point and the global objective function, and biases are caused by different local \sgd{} steps taken by clients. They propose \fednova{} to reduce the inconsistency by normalizing local updates with the number of local steps. The convergence analysis of \fednova{} assumes that all local objective functions have the same \(L\)-smoothness, which is also identical to the smoothness constant of the global objective function. However, in federated learning with highly heterogeneous datasets, smoothness constants \(L_k\) of local objective functions are very different across clients and from the one \(L_g\) of the global objective function. Although the assumption and proof still holds if taking \(L_g := \max (L_k)\) for all \(k \in [K]\), we argue that the inconsistency still exists in \fednova{} if only normalizing local updates with number of steps regardless of differences of \(L_k\)-smoothness constant of local objectives.

In one communication round, with the same numbers of local \sgd{} steps and a fixed learning rate \(\eta\), it is likely to happen that while objectives with small \(L\)-constant are still slowly converging, local objectives with large \(L\)-constants have converged in a few steps and extra steps are ineffective. In such cases, normalizing local updates with number of local \sgd{} steps implicitly over-weights updates from objectives with smaller \(L\)-constants when computing pseudo-gradients. Normalization of local updates to de-bias the pseudo-gradients is yet to be improved to take both different numbers of local steps and \(L\)-smoothness constants of local objectives into consideration. 

\section{\ours{}}\label{sec:adafedadam}
To address the drawbacks mentioned above, we propose \ours{} (Adaptive \fedadam{}) to make better use of accumulated local updates for fair federated learning (\eqref{eq:iffl}) with little efforts on fine-tuning. 
\subsection{Algorithm}

\begin{figure}[t]
% \centerline{\includegraphics[width=\textwidth]{figures/AdaFedAdam.pdf}}
\centering
\includegraphics[width=0.7\textwidth]{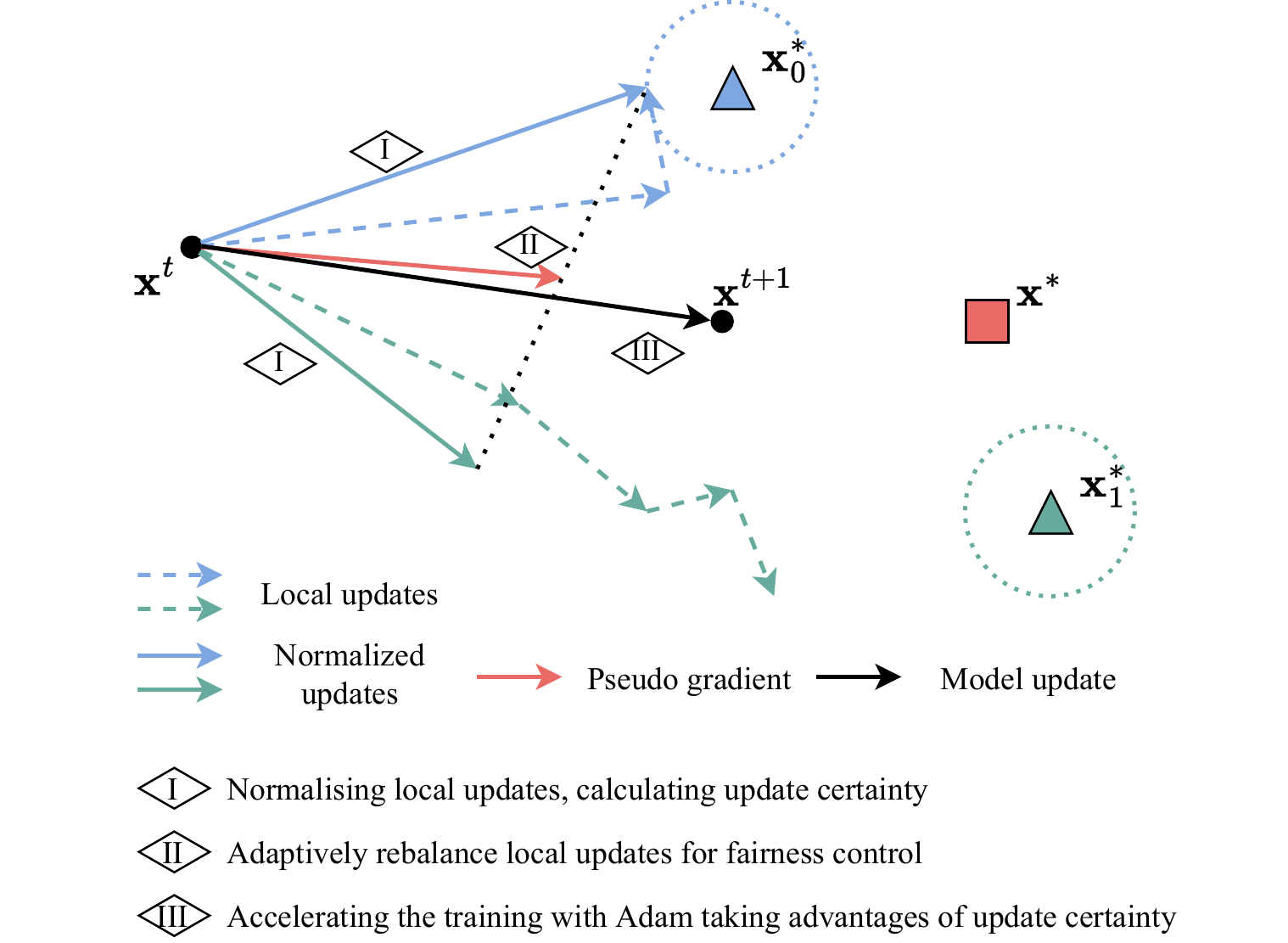}
\caption{Illustration of \ours{}: \(\mathbf{x}^t\), \(\mathbf{x}^*_k\) and \(\mathbf{x}^*\) denote the global model at round \(t\), the optima of local objective functions of client \(k\) and the optima of the global objective function, respectively. }
\label{fig:regu}
% \vspace{-3mm}
\end{figure}

The pseudo-code of the algorithm is presented as Algorithm \ref{alg:adafedadam} and Figure \ref{fig:regu} is an illustration of \ours{}. 

{\centering
\begin{minipage}{.7\linewidth}
\begin{algorithm}[H]
\small
\caption{Adaptive FedAdam}\label{alg:adafedadam}
\begin{algorithmic}
\Require initial model \(\mathbf{x}^0\), \(\eta\), \(\beta_1\), \(\beta_2\), \(\epsilon\)
\State \(m_{0} \gets 0\), \(v_{0} \gets 0\)
\State \(c_{0,m} \gets 1\), \(c_{0,v} \gets 1\)    \Comment Correction factor for \(m\) and \(v\)
% \State     \Comment Correction factor for \(v\)
\For{round \(t\) in \(\{0, 1, ... T-1\}\)}
    \State \(\mathbf{g}^t, C^t = \textbf{GetPseudoGradient}(\mathbf{x}^t)\)
    \State \(\beta_{t, 1} \gets \beta^{C^t}_1\), \(\beta_{t, 2} \gets \beta^{C^t}_2\), \(\eta_t \gets C^t\eta\)
    \Comment Adaptive momentum and stepsize
    % \State  \Comment Adaptive stepsize
    \State \(c_{t+1,m} \gets c_{t,m} \beta_{t,1}\), \(c_{t+1,v} \gets c_{t,v} \beta_{t,2}\) \Comment Update correction factors
    \State \(m_{t+1} \gets (1-\beta_{t,1})\mathbf{g}^t + \beta_{t,1}m_{t}\)
    \State \(v_{t+1} \gets (1-\beta_{t,2})\mathbf{g}^t \odot \mathbf{g}^t + \beta_{t,2}v_{t}\)
    \State \(\hat{m}_{t+1} \gets m_{t+1}/(1-c_{t+1, m})\)
    \State \(\hat{v}_{t+1} \gets v_{t+1}/(1-c_{t+1, v})\)
    \State \(\mathbf{x}^{t+1} \gets \mathbf{x}^{t} - \eta \hat{m}_{t+1}/(\sqrt{\hat{v}_{t+1}}+\epsilon)\)
\EndFor
\end{algorithmic}
\end{algorithm}
\end{minipage}
\par
}

{\centering
\begin{minipage}{.7\linewidth}
\begin{algorithm}[H]
\small
\caption{Pseudo-gradient calculation}
\begin{algorithmic}
\Require \(\alpha\)
\Procedure{GetPseudoGradient}{$\mathbf{x}$}
% \REQUIRE Initial global model \(\mathbf{x}^0\), \(\beta_{0,1}\), \(\beta_{0,2}\), \(\eta_{0}\)  and \(\eta_c\)
\State Broadcast \(\mathbf{x}\) to all clients
\For{client \(k\) in \(\{0, 1, ... K-1\}\) \textbf{parallel}} 
    \State Calculate \(F_k(\mathbf{x})\) and \(\nabla F_k(\mathbf{x})\)
    \State \(\mathbf{x}_k = \textbf{LocalSolver}(\mathbf{x}, \eta_k)\) \Comment Client training
    \State \(\mathbf{\Delta}_k \gets \mathbf{x}_k-\mathbf{x}\)
    \State \(\eta'_k \gets \frac{\|\mathbf{\Delta}_k\|_{2}}{\|\nabla F_k(\mathbf{x}^t)\|_{2}}\)
    \State \(\mathbf{U}_k \gets -\frac{\mathbf{\Delta}_k}{\eta'_k}\), \(C_k \gets \log (\eta'_k/\eta_k) + 1\)
    \Comment Normalization \& Certainty estimation
    \State \(I_k = F_k(\mathbf{x}) / F_k(\mathbf{x}^0)\)
    \State Report \((\mathbf{U}_k, C_k, I_k)\) to the server
\EndFor
\State \(\mathbf{g} \gets \frac{\sum S_k I_k^\alpha \mathbf{U}_k}{\sum S_k I_k^\alpha}\), \(C \gets \frac{\sum S_k I_k^\alpha C_k}{\sum S_k I_k^\alpha}\) \Comment Server aggregation
\State \Return \(\mathbf{g}\), \(C\)
\EndProcedure
\end{algorithmic}
\end{algorithm}
\end{minipage}
\par
}

\ours{} has 3 improvements over standard \fedadam{}: 
\paragraph{1. Normalization of local updates} Due to different \(L\)-smoothness constants of local objectives and local steps across participants, lengths of accumulated updates \(\mathbf{\Delta}_k\) are not at uniform scales and normalization of local updates is necessary as discussed in Section \ref{sec:ana_fedadam}. Natural scales for local updates are the \(\ell_2\)-norms of local gradients \(\|\nabla F_k(\mathbf{x}^t)\|_2\) on client \(k\). By normalizing \(\mathbf{\Delta}_k\) to the same \(\ell_2\)-norm of \(\|\nabla F_k(\mathbf{x}^t)\|_2\), a normalized update \(\mathbf{U}_k\) and a supportive factor \(\eta'_k\) are obtained. Intuitively, \(\mathbf{\Delta}_k\) can be seen as one update step following a "confident" update direction \(-\mathbf{U}_k\) with a large learning rate \(\eta'_k\) on the model \(\mathbf{x}^t\) given by client \(k\). The certainty of the direction \(\mathbf{U}^k\) is defined as \(C_k := \log(\eta'_k/\eta_k) + 1\) (\(\eta_k\) as the learning rate of the local solver), and the greater \(C_k\) is, the larger update can be made following \(\mathbf{U}_k\).  

\paragraph{2. Fairness control} Following the formulation of the loss function in fair federated learning in Section \ref{sec:background}, the pseudo-gradient \(\mathbf{g}^t\) of the global model \(\mathbf{x}^t\) is correspondingly the average of the normalized local updates with adaptive weights \(I^\alpha_k\), where \(I_k\) is the \textit{inverse training rate} and \(\alpha\) is the predefined hyperparameter for fairness control. The certainty of \(\mathbf{g}^t\) is given by the weighted average of local certainties \(C_k\) for all \(k\in[K]\). 

\paragraph{3. Adaptive hyperparameters for federated \adam{}}
Hyperparameters of \fedadam{} are adapted as follows to make better use of pseudo-gradients \(\mathbf{g}\) from accumulated updates: 
\begin{itemize}[nosep]
    \item  \(\beta_{t,1} \gets \beta_{1} ^ {C}\), \(\beta_{t,2} \gets \beta_{2} ^ {C}\): Adaptive \(\beta_{t,1}\) and \(\beta_{t,2}\) dynamically control the weight of the current update for the momentum estimation. \ours{} assigns more weight to more "certain" pseudo-gradients to update the average, and thus \(\beta_1\) and \(\beta_2\) are adapted exponentially following the form of exponentially weighted moving average.  
    \item \(\eta_{t} \gets C\eta\): The base stepsize \(\eta_t\) is adjusted based on the certainty of the pseudo-gradient \(C\) as well. Greater certainty enables larger \(\eta_t\) to construct larger \emph{trust regions} and vice versa. 
\end{itemize}

Theoretically, \ours{} ensures the following features: 
\begin{itemize} % [wide]
    \item \textbf{Fairness guarantee}: The fairness of the model has been formulated into the objective function in fair federated learning to be optimized together with the error with theoretical \((p, \alpha)\)\textit{-Proportional Fairness} (\cite{mo2000fair}). Also, the algorithm can be adapted to different fairness levels by adjusting \(\alpha\) in the problem formulation. 
    \item \textbf{Fine-tuning free:} The adaptivity of \ours{} derives from dynamic adjustment of hyperparameters for \adam{}. All initial hyperparameters of \ours{} can be chosen as the default values in the standard \adam{} for the centralized setting, and they are adaptively adjusted during the federated training process. 
    \item \textbf{Allowance for resource heterogeneity:} Thanks to the normalization of local updates, \ours{} allows arbitrary numbers of local steps, which could be caused by resource limitation of clients (also known as resource heterogeneity). 
    \item \textbf{Compatibility with arbitrary local solvers:} The normalization of local updates only relies on the \(\ell_2\)-norm of the local gradient estimation. Thus, any first-order optimizers are compatible with \ours{}. 
\end{itemize}
These features of \ours{} are empirically validated and discussed in Section \ref{sec:experiments}. 
\subsection{Convergence analysis for \ours{}}
The convergence guarantee of \ours{} for convex functions is proved as follows.  

\begin{theorem}[Convergence of \ours{}]\label{thrm:ours}
% \small
\textit{Assume the \(L\)-smooth convex loss function \(f(\mathbf{x})\) has bounded gradients \(\|\nabla\|_{\infty} \leq G\) for all \(\mathbf{x} \in \mathbb{R}^d\), and hyperparameters \(\epsilon\), \(\beta_{2,0}\) and \(\eta_0\) are chosen according to the following conditions: \(\eta_0 \leq \epsilon/2L\) and \(1-\beta_{2,0} \leq \epsilon^2/16G^2\).The pseudo-gradient \(\mathbf{g}_t\) at step \(t\) is given by Algorithm \ref{alg:adafedadam} with its certainty \(C^t\). The convergence guarantee of \ours{} is given by: }

\begin{equation}\label{eq:ours}
 \frac{1}{T} \sum_{t=1}^T \|\nabla f(\mathbf{x}^t)\|^2) \leq \frac{2(f(\mathbf{x}^1) - f(\mathbf{x}^*))(\sqrt{\beta_{2,0}}G+\epsilon)}{RC\eta_0} 
\end{equation}
where \(R := \min_t (\min_i | \frac{\mathbf{g}_{t,i}}{\nabla_{t,i}} |)\) for all \(i \in [d], t \in [T]\) and \(C := \min C^t\) for \(t \in [T]\). 
\end{theorem}

The proof of Theorem \ref{thrm:ours} is deferred to the \ref{subsec:proof2}. By normalizing local updates to the same \(\ell_2\)-norm of local gradients, the convergence of \ours{} can be guaranteed. When the client optimizers are fixed as \sgd{} and 1 step is performed locally, the federated training is identical to minibatch \adam{} and Theorem \ref{thrm:ours} gives the identical convergence guarantee of \adam{} (\cite{reddi2019convergence}). It should be noticed that Theorem \ref{thrm:ours} does not provide a tight bound for the convergence rate but only focuses on the convergence guarantee. Better empirical performance of \ours{} can be expected. 

\section{EXPERIMENTAL RESULTS}\label{sec:experiments}
\paragraph{Experimental setups}To validates the effectiveness and robustness of \ours{}, four federated setups are used: 1). Femnist setup: A multi-layer perceptron (MLP) network (\cite{pal1992multilayer}) for image classification on Federated EMNIST dataset (\cite{deng2012mnist}), proposed by \cite{caldas2018leaf} as a benchmark task for federated learning; 2). Cifar10 setup: VGG11 (\cite{simonyan2014very}) for image classification on Cifar10 dataset (\cite{krizhevsky2009learning}) partitioned by Dirichlet distribution \(\textbf{Dir}(0.05)\) for 16 clients; 3). Sent140 setup: A stacked-LSTM model (\cite{gers2000learning}) for sentiment analysis on the Text Dataset of Tweets (\cite{go2009twitter}); 4). Synthetic setup: A linear regression classifier for multi-class classification on a synthetic dataset (Synthetic), proposed by \cite{caldas2018leaf} as a challenging task for benchmarking federated algorithms. A summary of four setups are shown in Table \ref{tab:setups}. Details of the model architectures and experimental settings are available in \ref{subsec:exp_details}. 

\begin{table}[ht]
\small
  \caption{A summary of setups: the four setups cover different Federated Learning scenarios, non-IID types and task types. }
  \label{tab:setups}
  \centering
  \begin{tabular}{ l | l | l | l | l | l}
    \hline
    \header{Setup} & \header{\# Clients} & \header{Model} & \header{Scenario} & \header{Non-IID Type} & \header{Task Type}\\
    \hline
    Femnist     & 3500  &   MLP     & Cross Device      & Intrinsic  & Computer Vision\\
    Cifar10     & 16    &   CNN   & Cross Silo        & Dirichlet  & Computer Vision\\
    Sent140     & 697   &   LSTM    & Cross Device      & Intrinsic  & Natural Language Process\\
    Synthetic   & 100   &   Linear Model   & Cross Device/Silo & Synthetic  & Classification\\
    \hline
  \end{tabular}
  \vspace{-3mm}
\end{table}

\begin{figure*}[ht]
    \centering
    \includegraphics[width=0.33\textwidth]{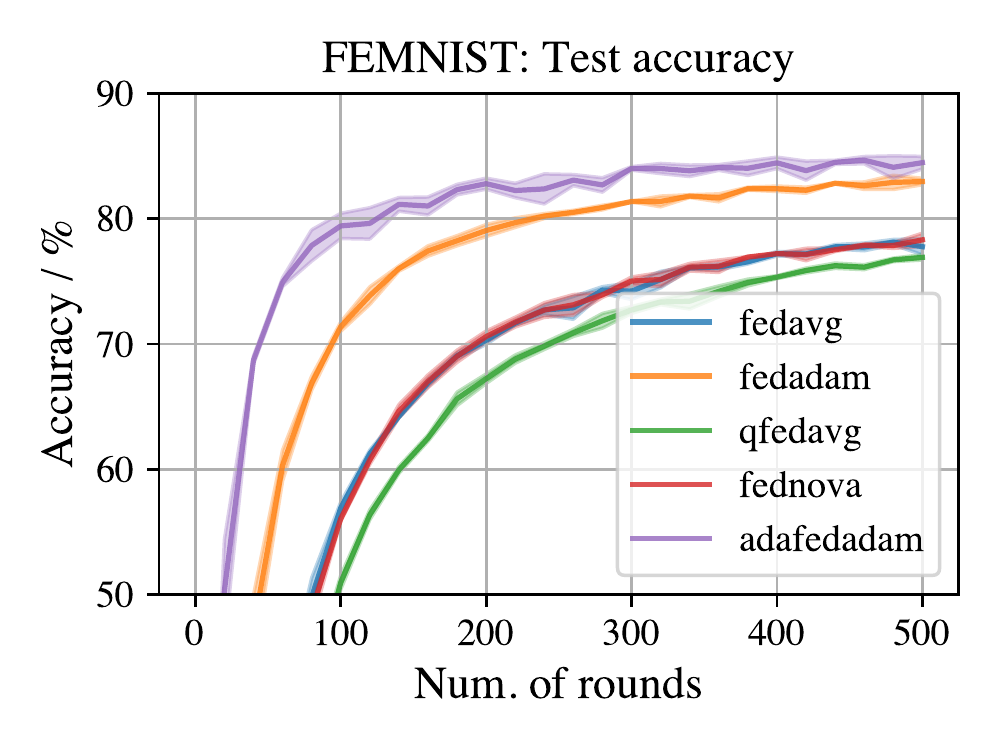}
    \includegraphics[width=0.33\textwidth]{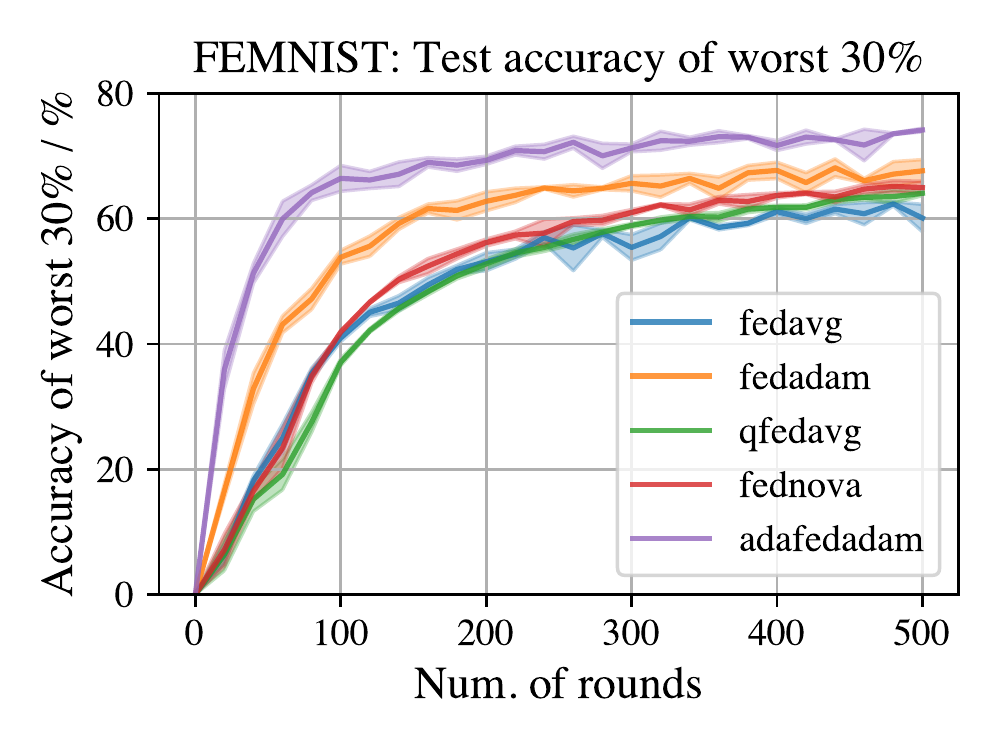}
    \includegraphics[width=0.33\textwidth]{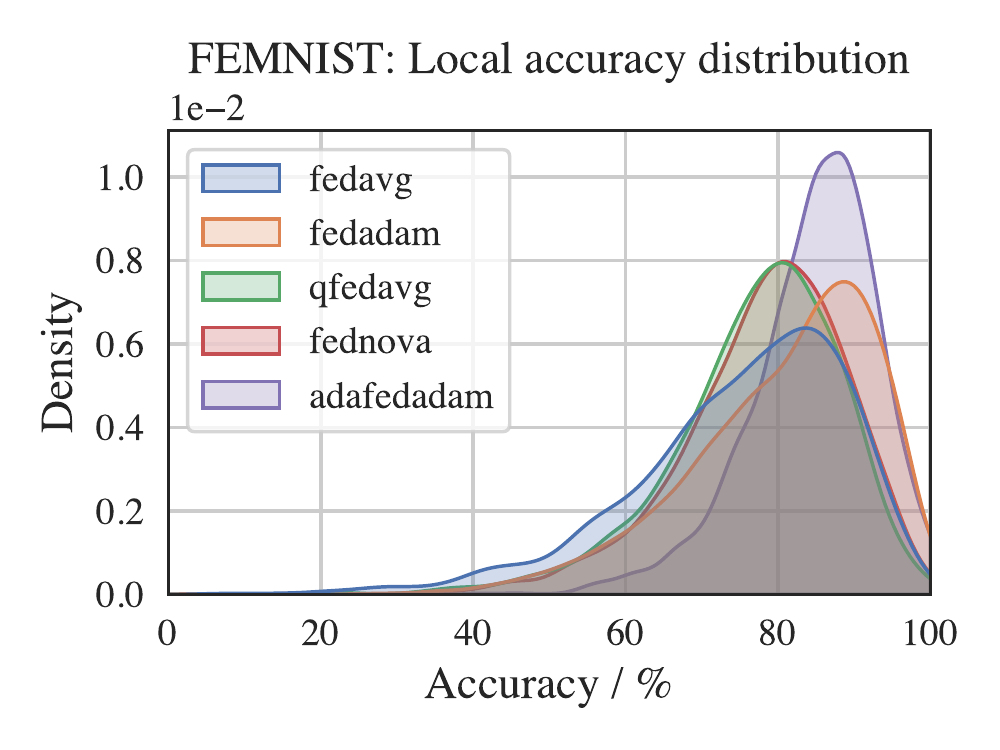}
    \includegraphics[width=0.33\textwidth]{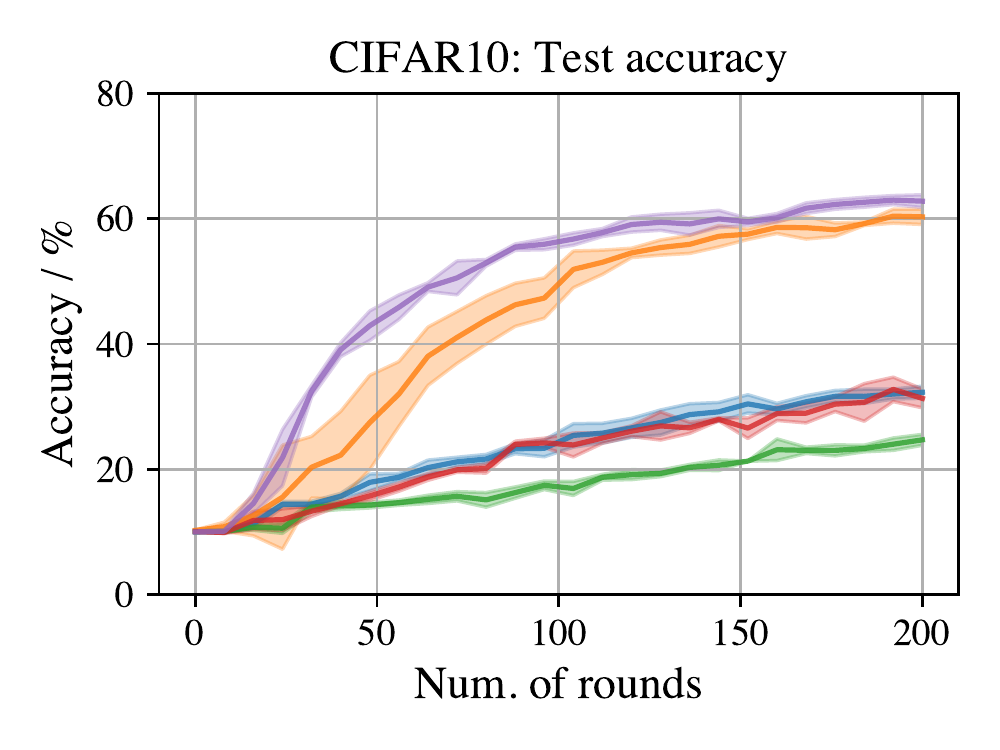}
    \includegraphics[width=0.33\textwidth]{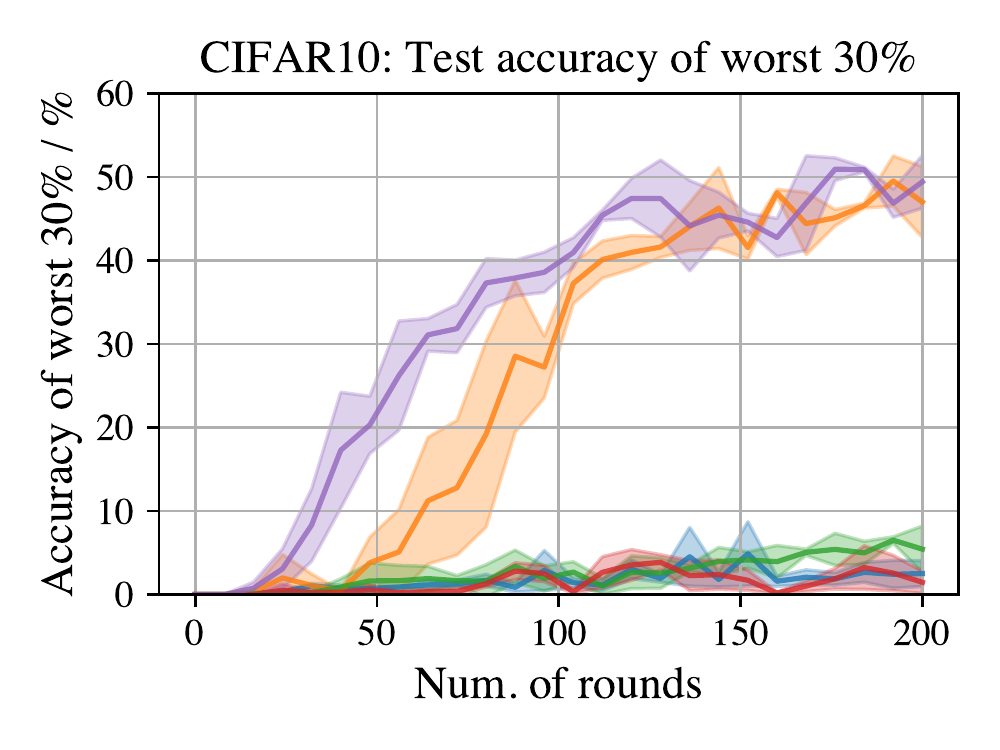}
    \includegraphics[width=0.33\textwidth]{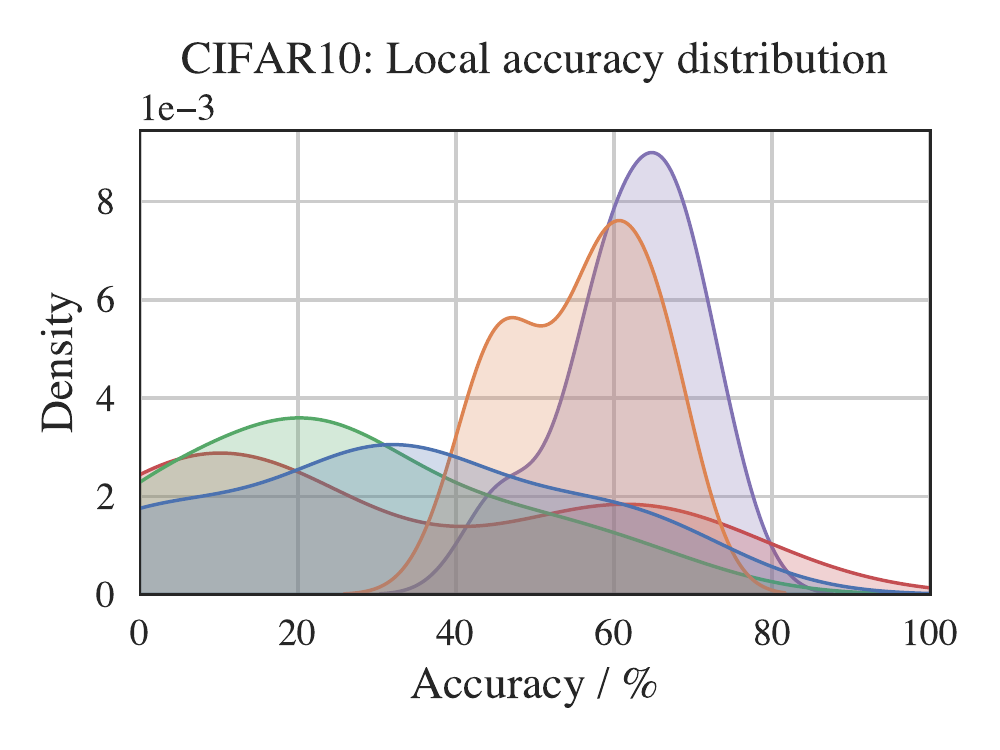}
    \includegraphics[width=0.33\textwidth]{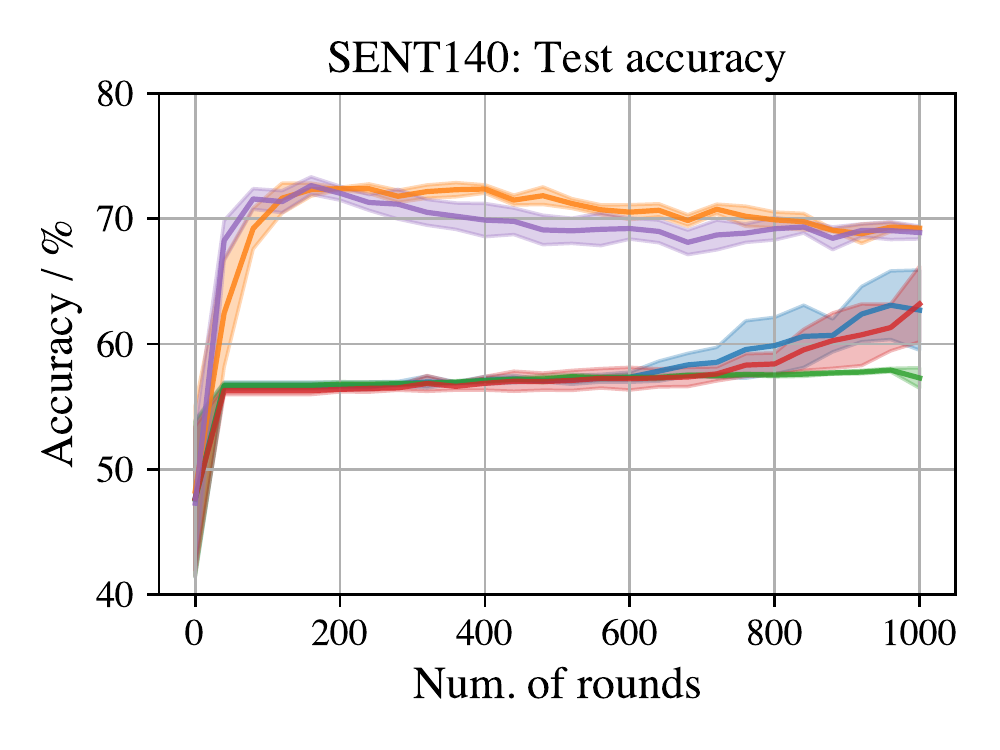}
    \includegraphics[width=0.33\textwidth]{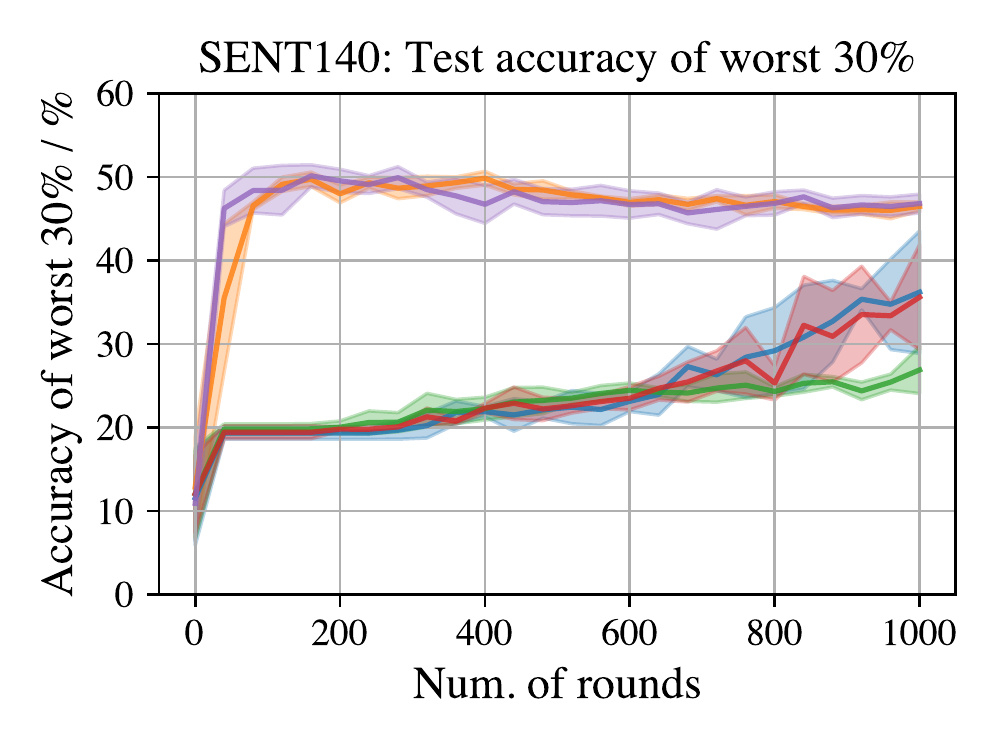}
    \includegraphics[width=0.33\textwidth]{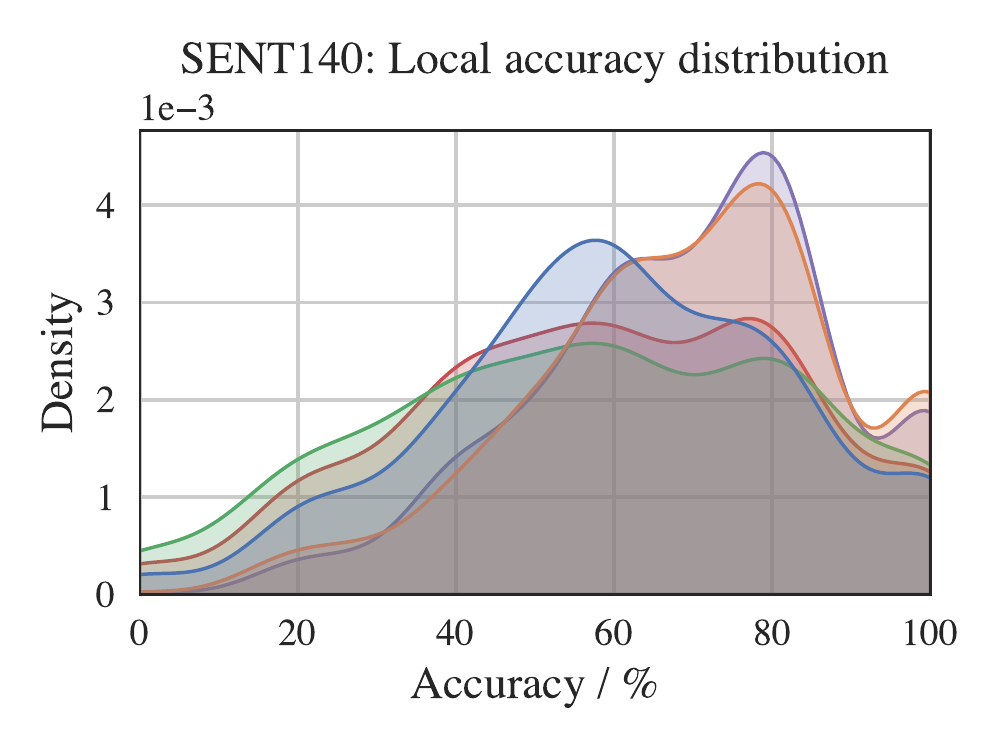}
    \includegraphics[width=0.33\textwidth]{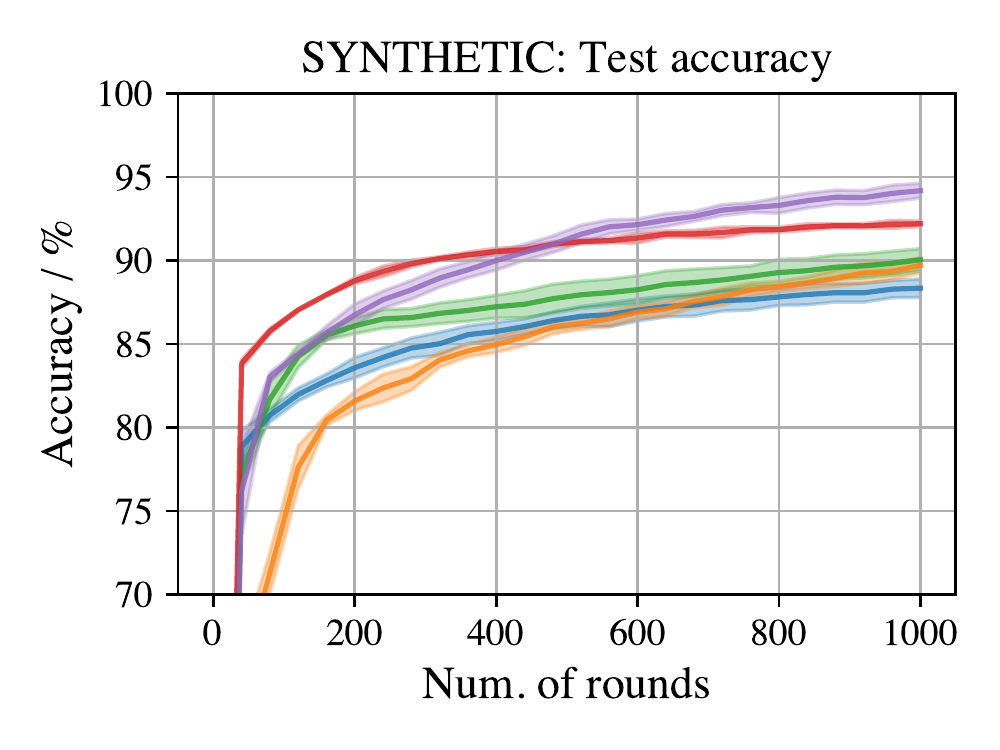}
    \includegraphics[width=0.33\textwidth]{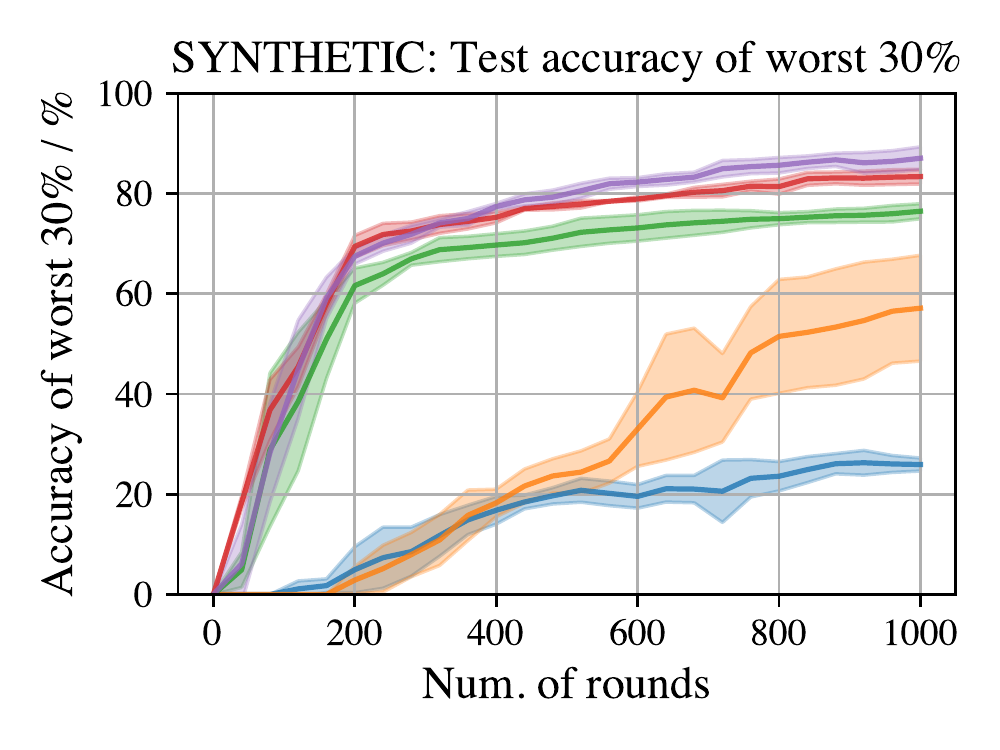}
    \includegraphics[width=0.33\textwidth]{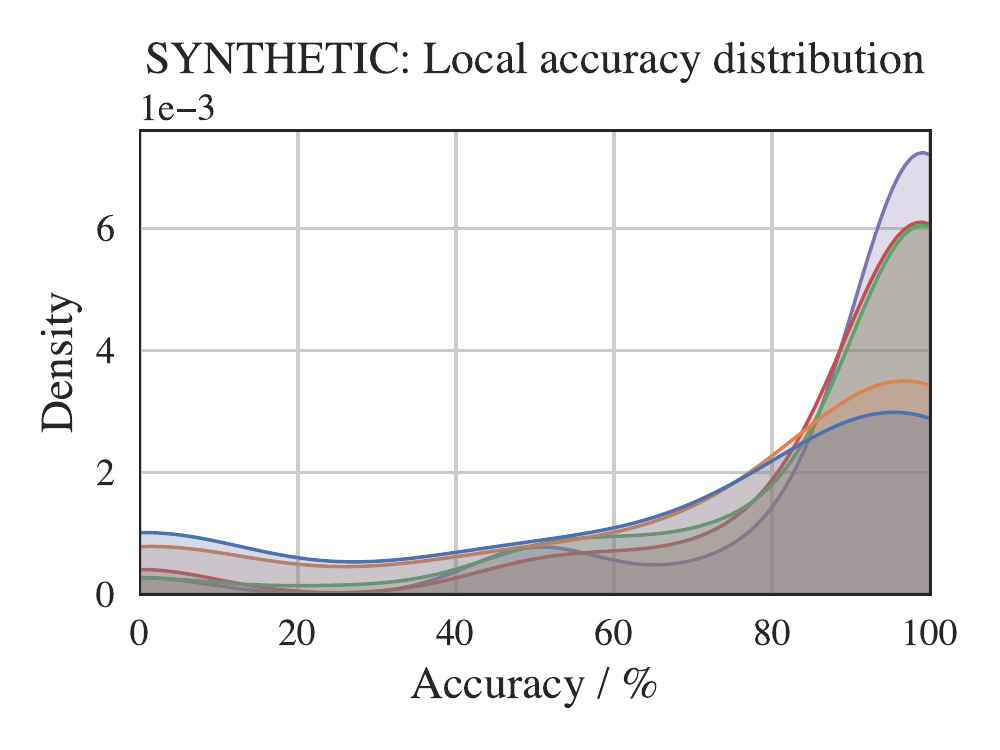}
    \caption{Metrics of local test accuracy during the training process: \fedavg{}, \fedadam{}, \fednova{}, \qfedavg{} and \ours{} on different setups. \textbf{Left: }Average of local test accuracy of participants. \textbf{Middle: }Average of local test accuracy of the worst 30\% participants. \textbf{Right: }Distribution of local test accuracy of participants. For all setups, \ours{} consistently outperforms baseline algorithms, with better model convergence and fairness among the participants. }
    \label{fig:training_curves}
\end{figure*}

\paragraph{Convergence \& Fairness}We benchmark \ours{} against \fedavg{}, \fedadam{}, \fednova{} and \qfedavg{} with \(q=1\). All hyperparameters of the optimizers are set as the default values in centralized settings. The fairness of the model is quantified by the standard deviation (STD) of local accuracy on clients and the average accuracy of the worst 30\% clients. The training curves are shown in Figure \ref{fig:training_curves} with numerical details in \ref{subsec:full_results}. Figure \ref{fig:training_curves} shows that \ours{} consistently converges faster than other algorithms, with better worst 30\% client performance for all setups. Distributions of local accuracy indicate that federated models trained with \ours{} provide the most uniform distributions of local accuracy for the participants. It is also noticeable that other federated algorithms without fine-tuning does not provide consistent performance in different setups with default hyperparameters, which validates the necessity of fine-tuning. In contrast, \ours{} provides the best and the most stable performances in all setups with default hyperparameters. To summarize, \ours{} is able to train federated models with fair performances among participants with better convergence without fine-tuning. 

\begin{figure*}[ht]
\centering
\includegraphics[width=0.33\textwidth]{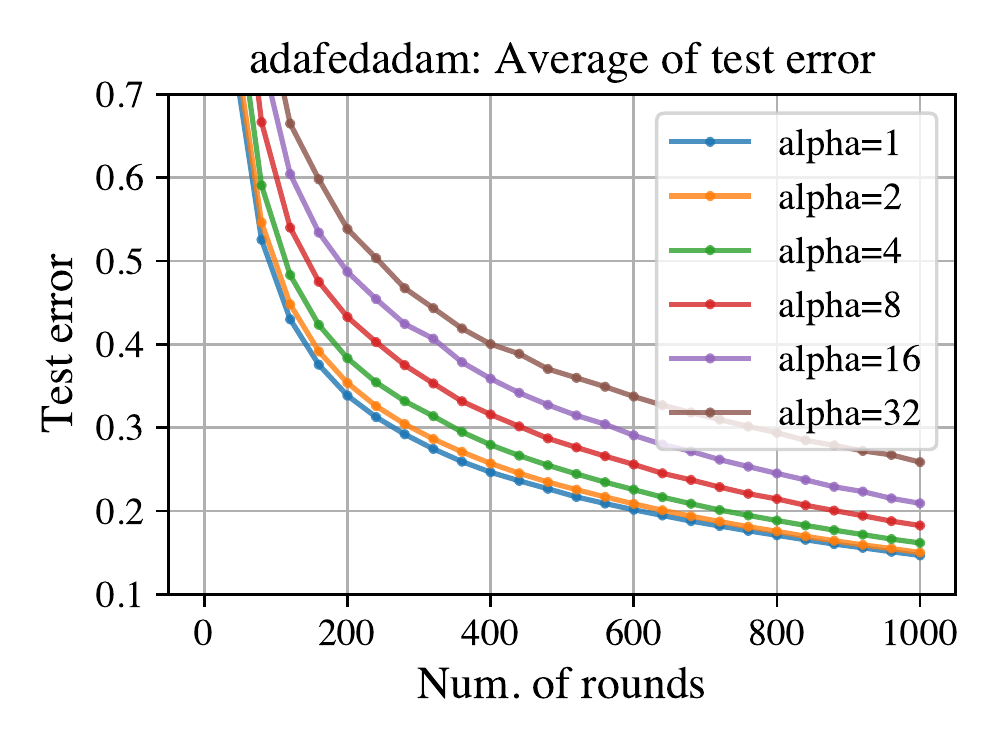}
\includegraphics[width=0.33\textwidth]{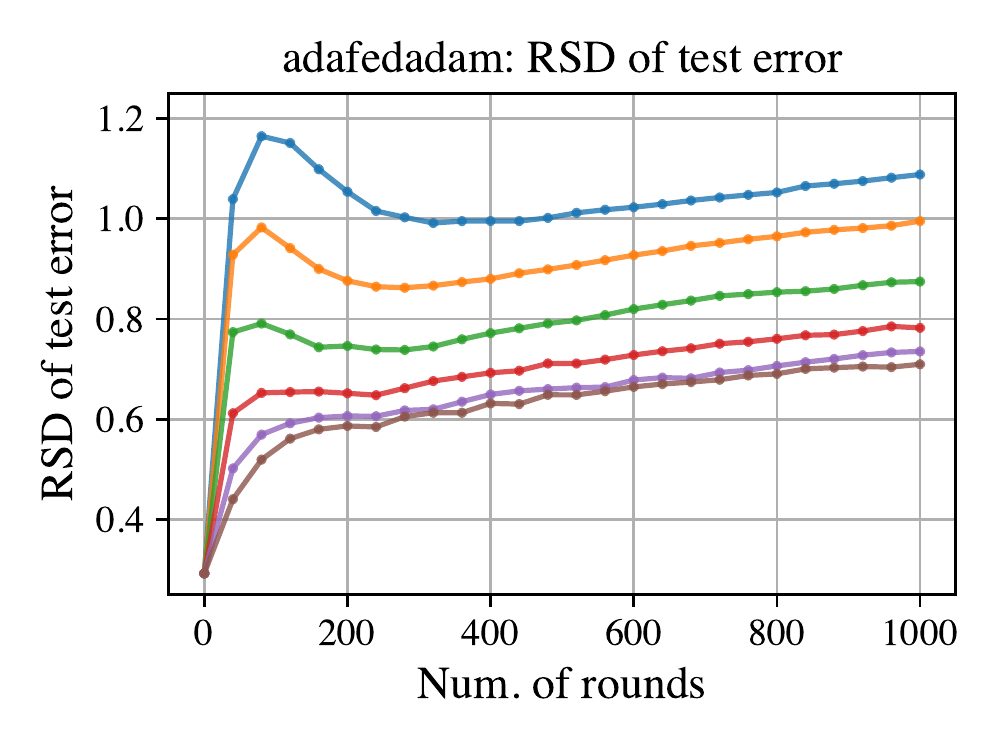}
\includegraphics[width=0.33\textwidth]{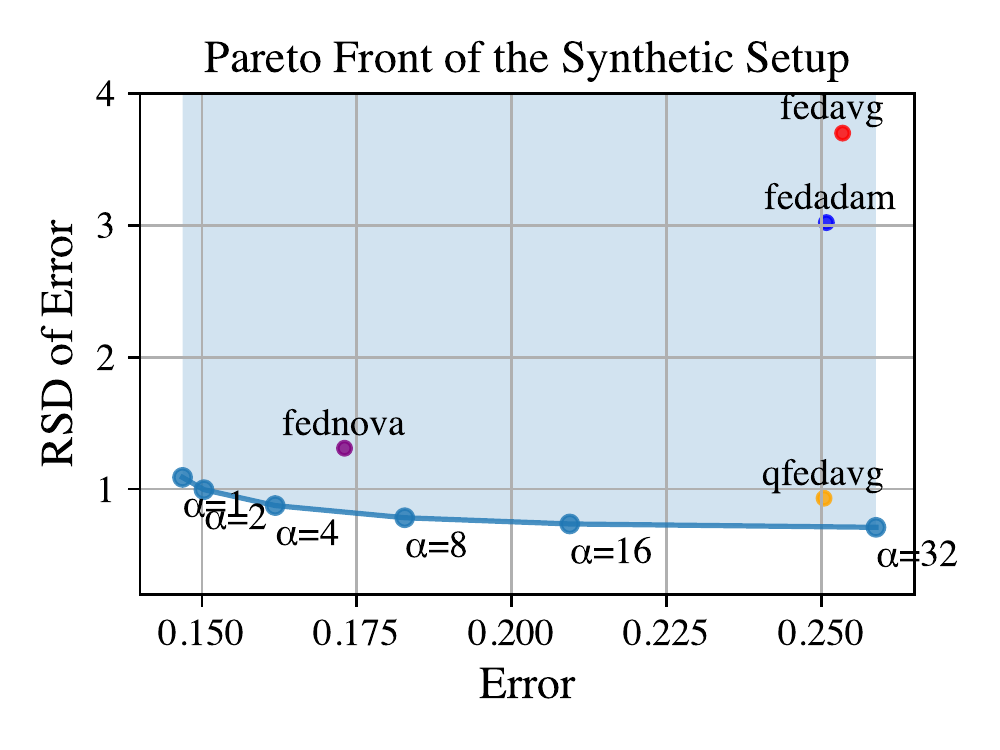}
\caption{\textbf{Left:} Training curves of \ours{} with different \(\alpha\) on the Synthetic setup. \textbf{Middle:} RSD of local error of \ours{} with different \(\alpha\) on the Synthetic setup. \textbf{Right:} Pareto Front of the Synthetic setup formed by \ours{} with different \(\alpha\). By adjusting values of \(\alpha\), the trade-off between the convergence and the fairness can be observed together with the suboptimality of other federated algorithms. }
\label{fig:pareto}
\end{figure*}

\paragraph{Choice of \(\alpha\)}Hyperparameter \(\alpha\) is to control the level of desired model fairness. By increasing \(\alpha\), models become more fair between clients at a cost of convergence. For each federated learning process, there exists a Pareto Front (\cite{ngatchou2005pareto}) for the trade-off. Taken the Synthetic setup as an example, the average and relative standard deviation (RSD) of local validation error during the training process and the formed Pareto Front is shown as Figure \ref{fig:pareto}. It is observed that with increase of \(\alpha\) from 1 to 4, the RSD of the local error decreases significantly with a slight decrease of the convergence. With \(\alpha>4\), the RSD of the local error does not reduce significantly but the convergence continues to decrease. By plotting the average and RDS of local error of models trained with \ours{} for different \(\alpha\) together with other federated algorithms, it can be observed that \fedavg{}, \fedadam{}, \fednova{} and \qfedavg{} are sub-optimal in the blue area in Figure \ref{fig:pareto}. By default, \(1\leq\alpha \leq4\) is enough to provide proper model fairness without losing much convergence. 

\paragraph{Robustness}Experiments to validate the robustness of \ours{} against resource heterogeneity and different levels of data heterogeneity are conducted with the Cifar10 setup. 

Robustness against resource heterogeneity is important for algorithms to be applied in real life. Due to the heterogeneity of clients' computing resources, the server cannot expect all participants perform requested number of local steps / epochs in each global round and thus, clients may perform arbitrary numbers of local steps on the global model in each communication round. To simulate settings of resource heterogeneity, time-varying numbers of local epochs are randomly sampled from a uniform distribution \(\mathcal{U}(1,3)\) in each communication round for each participant. The results are shonw in Table \ref{tab:rob_as}. With resource heterogeneity, \ours{} outperform other federated algorithms with higher average accuracy and more fairness. 

\begin{table}[ht]
\small
  \caption{Experimental results of federated algorithms against resource heterogeneity on the Cifar10 setup. Time-varying numbers of local epochs are sampled from a uniform distribution \(\mathcal{U}(1,3)\) in each communication round.}
  \label{tab:rob_as}
  \centering
  \begin{tabular}{l|lll}
    \hline
    \header{Algorithm} & \header{Avg.(\%)} & \header{STD.(\%)} & \header{Worst 30. (\%)} \\
    \hline
        \fedavg & 36.82 \textpm 1.45 & 21.32 \textpm 1.89 & 10.69 \textpm 3.29\\
        \fedadam & 54.57 \textpm 1.87 & 13.03 \textpm 2.53 & 40.32 \textpm 5.11\\
        \qfedavg & 27.36 \textpm 1.09 & 24.34 \textpm 1.35 & 2.67 \textpm 1.16\\
        \fednova & 38.03 \textpm 1.18 & 24.99 \textpm 2.15 & 6.24 \textpm 2.96\\
        \ours & 63.26 \textpm 1.41 & 8.64 \textpm 1.35  &45.07 \textpm 2.38\\

    \hline
  \end{tabular}
  \vspace{-3mm}
\end{table}

Robustness against different non-IID levels ensures the performance of an algorithm in various application cases. To simulate different non-IID levels, the Cifar10 dataset is partitioned by the Dirichlet distribution over labels with different concentration parameters \(\beta \in [0,1]\), denoted as \(\textbf{Dir}(\beta)\). A smaller value of \(\beta\) indicates larger level of data heterogeneity. The results are shown in Table \ref{tab:rob_dir}. With different levels of data heterogeneity, \ours{} is able to converge, and better performance and fairness are obtained in settings with less data heterogeneity, as expected. 

\begin{table}[H]
\small
  \caption{Experimental results of \ours{} in different levels of non-iid settings on the Cifar10 setup. }
  \label{tab:rob_dir}
  \centering
  \begin{tabular}{l|lll}
    \hline
    \header{Distribution} & \header{Avg.(\%)} & \header{STD.(\%)} & \header{Worst 30. (\%)} \\
    \hline
    \(\textbf{Dir}(0.05)\)  &62.81 \textpm 1.02 & 8.18 \textpm 1.33  &46.01 \textpm 2.23\\

    \(\textbf{Dir}(0.1)\)   &66.16 \textpm 1.13 & 6.58 \textpm 0.77 & 55.26 \textpm 2.83\\

    \(\textbf{Dir}(0.5)\)   & 71.43 \textpm 0.81 & 5.4 \textpm 0.22 & 64.93 \textpm 1.04\\
    
    \(\textbf{Dir}(1)\)    & 72.77 \textpm 0.44 & 3.05 \textpm 0.11 & 69.57 \textpm 0.54\\

    \hline
  \end{tabular}
  \vspace{-3mm}
\end{table}

\paragraph{Compatibility with local momentum} We also show that \ours{} is compatible with momentum-based local optimizers beside \sgd{}, which can further improve the model performance. Adaptive optimizers as client solvers (e.g. \adam{}) do not guarantee better performance over vanilla \sgd{} without synchronizing states of local optimizers, as discussed in \cite{yu2019linear,yuan2020federated}. There are reported algorithms to synchronize states of local optimizers and \ours{} is orthogonal and compatible with these algorithms. Full experimental results for different local solvers are deferred to \ref{subsec:full_results} 

\begin{table}[ht]
\small
  \caption{Experimental results of the \ours{} with different local optimizers on the Synthetic setup, including vanilla \sgd{}, \sgd{} with momentum and \sgd{} with Nesterov momentum. }
  \label{tab:local_solvers}
  \centering
  \begin{tabular}{l|lll}
    \hline
    \header{Local Optimizer} & \header{Avg.(\%)} & \header{STD.(\%)} & \header{Worst 30. (\%)} \\
    \hline
    Vanilla \sgd{}  & 94.18 \textpm 0.45 & 8.52 \textpm 0.37 & 87.07 \textpm 2.28\\

    \sgd{} w. Momen. & 97.19 \textpm 0.11 & 3.32 \textpm 0.02 & 93.41 \textpm 0.11\\

    \sgd{} w. Neste. Momen. & 97.27 \textpm 0.16 & 3.19 \textpm 0.21 & 94.19 \textpm 0.24\\

    \hline
  \end{tabular}
  \vspace{-3mm}
\end{table}

\section{Conclusion}\label{sec:conclusion}
In this work, we formulated federated learning as a dynamic multi-objective optimization problem by adjusting the weights of local objectives to achieve fair model performance among the participants. To solve the problem efficiently, we presented \ours{}, which reduces biases in \fedadam{} and accelerates the training of fair federated learning with minor extra efforts in fine-tuning. Empirically we validated the efficiency and fairness of \ours{} and verified its Pareto optimality compared with other federated learning algorithms. Further, we demonstrated the robustness of \ours{} against resource heterogeneity and different levels of data heterogeneity. We have also shown the compatibility of \ours{} with other local optimizers. Future directions include testing \ours{} in real-world geographically distributed setups for both cross-silo and cross-device settings with production grade open source frameworks(\cite{yang2019federated, ekmefjord2022scalable}). 

% \section*{Acknowledgments}
% This was was supported in part by......

%Bibliography
% \newpage
\printbibliography[heading=subbibliography]

\newpage
\appendix
\section{Proof for Theorems}\label{sec:proof}
\subsection{Proof for Theorem 1}\label{subsec:proof1}
In this section we provide the proof for Theorem 1. \\

\textbf{Lemma: (Path length bound for Stochastic Gradient Descent)}

With same assumptions for function \(f(\mathbf{x})\) in Theorem 1, if the SGD iterates with learning rate \(\eta\) exhibit approximately linear convergence with constants \((A, c)\) for \(N\) steps, then the path length \(\mathcal{L}_N := \sum_0^N \|\mathbf{x}^{n+1} - \mathbf{x}^{n} \|_2\) is bounded as:
\begin{align*}
    \mathcal{L}_N 
    \leq & \|\mathbf{x}^0 - \mathbf{x}^* \|_2 \sum_0^N (1-c)^n\eta AL \\
    \leq & \|\mathbf{x}^0 - \mathbf{x}^* \|_2 \frac{1-(1-c)^N}{c}AL
\end{align*}

The proof of the lemma can be referred to \cite{gupta2021path}. \\

Here we analyze the convergence with no momentum (\(\beta_1=0\)), and the result can be extended to general cases (\cite{zaheer2018adaptive}). 

To simplify the notation, we denote \(\nabla_{t,i}\) as the \(i\)th element of the gradient of model \(\nabla f(\mathbf{x}_t)\) at round \(t\), and \(\Delta_{t,i}\) for the \(i\)th element of \(\Delta_t\). The path length of \sgd{} updates for at step \(t\) is denoted as \(\mathcal{L}_N^t\)

Recall that the update rule of \naccadam{} is given by 
\begin{align*}
    \mathbf{x}_{t+1} = \mathbf{x}_{t} - \eta \frac{\Delta_t}{\sqrt{v_t}+\epsilon}
\end{align*}
for all \(i \in [d]\). Let \(R_t := \min | \frac{\Delta_{t,i}}{\nabla_{t,i}} |\) for \(i\in[d]\) and \(\mathcal{P}_N^t := \frac{\|\Delta_t\|_2}{\|\nabla f(\mathbf{x}^t)\|_2}\). L-smoothness of function \(f(\mathbf{x})\) guarantees that

\begin{align*}
    f(\mathbf{x}_{t+1}) &\leq f(\mathbf{x}_t) + \langle \nabla_t, \mathbf{x}_{t+1} - \mathbf{x}_t\rangle + \frac{L}{2}\|\mathbf{x}_{t+1} - \mathbf{x}_t \|^2_2\\
    &= f(\mathbf{x}_{t}) - \eta\sum_{i=1}^d (\nabla_{t,i} \cdot \frac{\Delta_{t,i}}{\sqrt{v_{t,i}}+\epsilon}) + \frac{L\eta^2}{2}\sum_{i=1}^d \frac{\Delta^2_{t,i}}{(\sqrt{v_{t,i}}+\epsilon)^2}\\
    &= f(\mathbf{x}_t) - \eta\sum_{i=1}^d(\nabla_{t,i} \cdot (\frac{\Delta_{t,i}}{\sqrt{v_{t,i}}+\epsilon} - \frac{R_t\nabla_{t,i}}{\sqrt{\beta_2v_{t-1,i}}+\epsilon} \\
    & + \frac{R_t\nabla_{t,i}}{\sqrt{\beta_2v_{t-1,i}}+\epsilon})) + \frac{L\eta^2}{2}\sum_{i=1}^d \frac{\Delta^2_{t,i}}{(\sqrt{v_{t,i}}+\epsilon)^2}\\
    &\leq f(\mathbf{x}_t) - R_t\eta\sum_{i=1}^d\frac{\nabla^2_{t,i} }{\sqrt{\beta_2v_{t-1, i}}+\epsilon} \\
    & + \eta\sum_{i=1}^d\nabla_{t,i} \underbrace{|\frac{\Delta_{t,i}}{\sqrt{v_{t,i}}+\epsilon} - \frac{R_t\nabla_{t,i}}{\sqrt{\beta_2v_{t-1,i}}+\epsilon} |}_{T} \\
    & + \frac{L\eta^2}{2}\sum_{i=1}^d \frac{\Delta^2_{t,i}}{(\sqrt{v_{t,i}}+\epsilon)^2}
\end{align*}

\(T\) is bounded by
\begin{align*}
    T &= | \frac{\Delta_{t,i}}{\sqrt{v_{t,i}}+\epsilon} - \frac{R_t\nabla_{t,i}}{\sqrt{\beta_2v_{t-1,i}}+\epsilon} |\\
        &\leq | \frac{\Delta_{t,i}}{\sqrt{v_{t,i}}+\epsilon} - \frac{\Delta_{t,i}}{\sqrt{\beta_2v_{t-1,i}}+\epsilon} | \\
        &\leq |\Delta_{t,i} | \cdot |\frac{1}{\sqrt{v_{t,i}}+\epsilon} - \frac{1}{\sqrt{\beta_2v_{t-1,i}}+\epsilon} | \\
        &= \frac{|\Delta_{t,i} |} {(\sqrt{v_{t,i}} + \epsilon)(\sqrt{\beta_2 v_{t-1,i}} +\epsilon)} \cdot  \frac{(1-\beta_2)\Delta^2_{t,i}}{\sqrt{v_{t,i}} + \sqrt{\beta_2v_{t-1,i}}} \\
        &\leq \frac{1}{(\sqrt{v_{t,i}}+\epsilon)(\sqrt{\beta_2v_{t-1,i}}+\epsilon)} \cdot \sqrt{1-\beta_2}\Delta^2_{t,i} \\ 
        &\leq \frac{\sqrt{1-\beta_2}\Delta^2_{t,i}}{(\sqrt{\beta_2v_{t-1,i}} + \epsilon)\epsilon}
\end{align*}

With the bound above and \(\|\nabla f(\mathbf{x}_t)\|_\infty \leq G\) for all \(i \in [d]\), we have following

\begin{align*}
    f(\mathbf{x}_{t+1}) &\leq f(\mathbf{x}_t) - R_t\eta\sum_{i=1}^d\frac{\nabla^2_{t,i}}{\sqrt{\beta_2v_{t-1, i}}+\epsilon} \\
    & + \frac{\eta G\sqrt{1-\beta_2}}{\epsilon} \sum_{i=1}^{d}\frac{\Delta^2_{t,i}}{\sqrt{\beta_2v_{t-1,i}}+\epsilon} + \frac{L\eta^2}{2\epsilon}\sum_{i=1}^d \frac{\Delta^2_{t,i}}{\sqrt{v_{t,i}}+\epsilon} \\
    &\leq f(\mathbf{x}_t) - \eta R_t \sum_{i=1}^d\frac{\nabla^2_{t,i}}{\sqrt{\beta_2v_{t-1, i}}+\epsilon} \\
    & + \frac{\mathcal{P}^t_N\eta G\sqrt{1-\beta_2}}{\epsilon} \sum_{i=1}^{d}\frac{\nabla^2_{t,i}}{\sqrt{\beta_2v_{t-1,i}}+\epsilon} +
    \frac{\mathcal{P}^t_N L\eta^2}{2\epsilon}\sum_{i=1}^d \frac{\nabla^2_{t,i}}{\sqrt{\beta_2v_{t-1,i}}+\epsilon}
\end{align*}

From the parameters \(\eta, \epsilon\) and \(\beta\) stated in \adam{}, \(L\eta/2\epsilon \leq 1/4\) and \(G\sqrt{1-\beta_2}/\epsilon \leq 1/4\) hold. Using the inequality conditions and let \(V_t := \frac{\|\Delta_t\|_2}{\|\nabla f(\mathbf{x}_t)\|_2}\), we have

\begin{align*}
    f(\mathbf{x}_{t+1}) &\leq f(\mathbf{x}_t) - (R_t - \frac{\mathcal{P}^t_N}{2})\eta \sum_{i=1}^d \frac{\nabla^2_{t,i}}{\sqrt{\beta_2v_{t-1,i}}+\epsilon} \\
    & \leq f(\mathbf{x}_t) - (\frac{R_t}{\mathcal{P}^t_N} - \frac{1}{2}) \frac{\eta}{\sqrt{\beta_2}G+\epsilon} \|\nabla f(\mathbf{x}_t)\|^2 
\end{align*}

Using a telescope sum and rearranging the inequality, we have
\begin{align*}
 \frac{\eta}{\sqrt{\beta_2}G+\epsilon} \sum_{t=1}^T (\frac{R_t}{\mathcal{P}^t_N} - \frac{1}{2}) \|\nabla f(\mathbf{x}_t)\|^2) \leq f(\mathbf{x}_1) - f(\mathbf{x}_{t+1})
\end{align*}

Due to the fact that \(0\leq R_t \leq N\) for all \(t\) and \(f(\mathbf{x}^*) \leq f(\mathbf{x}_{t+1})\), in the case where \(R_t \leq \mathcal{P}^t_N/2\), the algorithm does not converge. 

With \( \|\Delta_t\|_2 \leq \eta_s \mathcal{L}_N^t\) and \(\nabla f(\mathbf{x}^t) = \eta_s \mathcal{L}_1^t\), we have \(\mathcal{P}^t_N =\frac{\|\Delta_t\|_2}{\eta_s \mathcal{L}_1^t} \leq \frac{\mathcal{L}_N}{\mathcal{L}_1} \leq \frac{1-(1-c)^N}{c}\) for all \(t < T\). In the best case where \(R_t = N\), the convergence rate can be derived as follows: 

\begin{align*}
 \frac{1}{T} \sum_{t=1}^T \|\nabla f(\mathbf{x}_t)\|^2 &\leq \frac{f(\mathbf{x}_1) - f(\mathbf{x}^*)(\sqrt{\beta_2}G+\epsilon)}{(\frac{Nc}{1-(1-c)^N} - \frac{1}{2})\eta T} 
%  & = \mathcal{O}(\frac{N}{T})
\end{align*}

When \(N=1\), the convergence rate of \naccadam{} is the same as \adam{} (\cite{zaheer2018adaptive}). 

\subsection{Proof for Theorem 2}\label{subsec:proof2}
In this section we provide the proof for Theorem 2. We analyze the convergence with no momentum (\(\beta_1=0\)) and \(\alpha=0\) here. Similar to the proof for Theorem 1, the convergence analysis can be extended to general cases. The notation in the proof follows \ref{subsec:proof1}. In \ours{}, \(\mathbf{g}_t\) is given by \(\mathbf{g}_t := \frac{\sum S_k \mathbf{U}_k}{\sum S_k}\) where \(\mathbf{U}_k\) is the normalized local update given by client \(k\) in round \(t\) with its certainty \(C_k\) (i.e. \(\|\mathbf{U}_k\|_2\ = \|\nabla_k\|_2\) and  \(\Delta_{k} = 0.01 C_k \mathbf{U}_k\)). The certainty of \(\mathbf{g}_t\) is given by \(C_t := \sqrt{\frac{\sum S_k C_k}{\sum S_k}}\). 

Recall that the update rule of \ours{} is given by 
\begin{align*}
    \mathbf{x}_{t+1} = \mathbf{x}_{t} - (\log C_t+1)\eta_0 \frac{\mathbf{g}_t}{\sqrt{v_t}+\epsilon}
\end{align*}
for all \(i \in [d]\). Let \(R_t := \min \| \frac{\mathbf{g}_{t,i}}{\nabla f(\mathbf{x}_t)_{i}} \|\) for \(i\in[d]\). L-smoothness of the function \(f(\mathbf{x})\) guarantees that
\begin{align*}
    f(\mathbf{x}_{t+1}) 
    \leq & f(\mathbf{x}_t) + \langle \nabla_t, \mathbf{x}_{t+1} - \mathbf{x}_t\rangle + \frac{L}{2}\|\mathbf{x}_{t+1} - \mathbf{x}_t \|^2_2 \\
    =& f(\mathbf{x}_{t}) - (\log C_t+1)\eta_0\sum_{i=1}^d (\nabla_{t,i} \cdot \frac{\mathbf{g}_{t,i}}{\sqrt{v_{t,i}}+\epsilon}) + \frac{L((\log C_t+1)\eta_0)^2}{2}\sum_{i=1}^d \frac{\mathbf{g}^2_{t,i}}{(\sqrt{v_{t,i}}+\epsilon)^2}\\
    =& f(\mathbf{x}_t) - (\log C_t+1)\eta_0\sum_{i=1}^d(\nabla_{t,i} \cdot (\frac{\mathbf{g}_{t,i}}{\sqrt{v_{t,i}}+\epsilon} - \frac{R_t\nabla_{t,i}}{\sqrt{\beta_{2,t}v_{t-1,i}}+\epsilon} + \frac{R_t\nabla_{t,i}}{\sqrt{\beta_{2,t}v_{t-1,i}}+\epsilon})) \\ 
    &+ \frac{L((\log C_t+1)\eta_0)^2}{2}\sum_{i=1}^d \frac{\mathbf{g}^2_{t,i}}{(\sqrt{v_{t,i}}+\epsilon)^2}\\
    \leq& f(\mathbf{x}_t) - R_t(\log C_t+1)\eta_0\sum_{i=1}^d\frac{\nabla^2_{t,i} }{\sqrt{\beta_{2,t}v_{t-1, i}}+\epsilon} \\
    &+ (\log C_t+1)\eta_0\sum_{i=1}^d \nabla_{t,i} \underbrace{ | \frac{\mathbf{g}_{t,i}}{\sqrt{v_{t,i}}+\epsilon} - \frac{R_t\nabla_{t,i}}{\sqrt{\beta_{2,t}v_{t-1,i}}+\epsilon} |}_{T} \\
    &+ \frac{L((\log C_t+1)\eta_0)^2}{2}\sum_{i=1}^d \frac{\mathbf{g}^2_{t,i}}{(\sqrt{v_{t,i}}+\epsilon)^2}
\end{align*}

\(T\) is bounded by
\begin{align*}
    T &= | \frac{\mathbf{g}_{t,i}}{\sqrt{v_{t,i}}+\epsilon} -      \frac{R_t\nabla_{t,i}}{\sqrt{\beta_{2,t}v_{t-1,i}}+\epsilon} | \\
        &\leq | \frac{\mathbf{g}_{t,i}}{\sqrt{v_{t,i}}+\epsilon} - \frac{\mathbf{g}_{t,i}}{\sqrt{\beta_{2,t}v_{t-1,i}}+\epsilon} | \\
        &\leq |\mathbf{g}_{t,i} | \cdot |\frac{1}{\sqrt{v_{t,i}}+\epsilon} - \frac{1}{\sqrt{\beta_{2,t}v_{t-1,i}}+\epsilon} | \\
        &= \frac{|\mathbf{g}_{t,i} |} {(\sqrt{v_{t,i}} + \epsilon)(\sqrt{\beta_{2,t} v_{t-1,i}} +\epsilon)} \cdot  \frac{(1-\beta_{2,t})\mathbf{g}^2_{t,i}}{\sqrt{v_{t,i}} + \sqrt{\beta_{2,t}v_{t-1,i}}} \\
        &\leq \frac{1}{(\sqrt{v_{t,i}}+\epsilon)(\sqrt{\beta_{2,t}v_{t-1,i}}+\epsilon)} \cdot \sqrt{1-\beta_{2,t}}\mathbf{g}^2_{t,i} \\ 
        &\leq \frac{\sqrt{1-\beta_{2,t}}\mathbf{g}^2_{t,i}}{(\sqrt{\beta_{2,t}v_{t-1,i}} + \epsilon)\epsilon}
\end{align*}

With the bound above and \(\|\nabla f(\mathbf{x}_t)\|_\infty \leq G\), we have following
\begin{align*}
    f(\mathbf{x}_{t+1}) \leq 
    & f(\mathbf{x}_t) - R_t(\log C_t+1)\eta_0\sum_{i=1}^d\frac{\nabla^2_{t,i}}{\sqrt{\beta_{2,t}v_{t-1, i}}+\epsilon}  \\
    &+ \frac{(\log C_t+1)\eta_0 G\sqrt{1-\beta_{2,t}}}{\epsilon} \sum_{i=1}^{d}\frac{\mathbf{g}^2_{t,i}}{\sqrt{\beta_{2,t}v_{t-1,i}}+\epsilon} \\
    &+ \frac{L((\log C_t+1)\eta_0)^2}{2\epsilon}\sum_{i=1}^d \frac{\mathbf{g}^2_{t,i}}{\sqrt{v_{t,i}}+\epsilon} \\
    \leq & f(\mathbf{x}_t) - (\log C_t+1)\eta_0 R_t \sum_{i=1}^d\frac{\nabla^2_{t,i}}{\sqrt{\beta_{2,t}v_{t-1, i}}+\epsilon} \\
    &+ \frac{(\log C_t+1)\eta_0 G\sqrt{1-\beta_{2,t}}}{\epsilon } \sum_{i=1}^{d}\frac{\nabla^2_{t,i}}{\sqrt{\beta_{2,t}v_{t-1,i}}+\epsilon} \\
    &+ \frac{L((\log C_t+1)\eta_0)^2}{2\epsilon}\sum_{i=1}^d \frac{\nabla^2_{t,i}}{\sqrt{\beta_{2,t}v_{t-1,i}}+\epsilon}
\end{align*}

From the parameters \(\eta, \epsilon\) and \(\beta\) stated in \adam{}, \(L\eta_0/2\epsilon \leq 1/4\) and \(G\sqrt{1-\beta_{2,0}}/\epsilon \leq 1/4\) hold. The inequality \(G\sqrt{1-\beta^{C_t}_{2,0}}/\eta \leq (\log C_t+1)/4\) holds if \(\beta_{2,0} \geq \log^2 2 \approx 0.520\) and \(C_t \geq 1\), which is true since \(\beta_{2,0}\) is close to 1 with the default value \(\beta_{2,0} = 0.999\) and \(C_t \geq 1\). Using the inequality conditions, we have
\begin{align*}
    f(\mathbf{x}_{t+1}) &\leq f(\mathbf{x}_t) - (R_t - \frac{(\log C_t+1)}{2})(\log C_t+1)\eta_0 \sum_{i=1}^d \frac{\nabla^2_{t,i}}{\sqrt{\beta_{2,t}v_{t-1,i}}+\epsilon} \\
    & \leq f(\mathbf{x}_t) - \frac{R_t}{2} \frac{(\log C_t+1)\eta_0}{\sqrt{\beta_{2,0}}G+\epsilon} \|\nabla f(\mathbf{x}_t)\|^2 
\end{align*}

The second inequality is due to the fact that \(R_t \geq C_t > (\log C_t+1)\) and \(\beta_{2,t} \leq \beta_{2,0}\) if \(C_t \geq 1\). Using a telescope sum and rearranging the inequality, we have
\begin{align*}
 \frac{\eta_0}{\sqrt{\beta_{2,0}}G+\epsilon} \sum_{t=1}^T R_t(\log C_t+1) \|\nabla f(\mathbf{x}_t)\|^2 \leq f(\mathbf{x}_1) - f(\mathbf{x}_{t+1})
\end{align*}

Let \(R := \min R_t\) and \(C := \min C_t\) for all \(t \in [T]\), by rearranging the inequality, we obtain

\begin{align*}
 \frac{1}{T} \sum_{t=1}^T \|\nabla f(\mathbf{x}_t)\|^2 &\leq \frac{2(f(\mathbf{x}_1) - f(\mathbf{x}^*))(\sqrt{\beta_{2,0}}G+\epsilon)}{R(\log C+1)\eta_0} 
%  & = \mathcal{O}(\frac{N}{T})
\end{align*}

\newpage
% \appendix

\section{PSEUDO CODES FOR ALGORITHMS}\label{sec:pseudocode}
Pseudo codes for \adam{}, \naccadam{} and minibatch \sgd{} are given as Algorithm \ref{alg:adam} and Algorithm \ref{alg:sgd}. 

\begin{algorithm}
\caption{\textcolor{red}{\adam{}} and \textcolor{blue}{\naccadam{}}\label{alg:adam}}
\begin{algorithmic}
\Require model weights $\mathbf{x}^0$, stepsize $\eta$, $\beta_1$, $\beta_2$, $\epsilon$ for both \textcolor{red}{\adam{}} and \textcolor{blue}{\naccadam{}}, $\eta_s$ and $N$ for \textcolor{blue}{\naccadam{}}
\State $m_0 \gets 0$
\State $v_0 \gets 0$
\For{step $t$ in $\{0, 1, ... T-1\}$}
    \State \textcolor{red}{\adam{}}: $\Delta_t = \nabla_{\zeta \sim \mathcal{D}} F(\mathbf{x}^{t})$
    \State \textcolor{blue}{\naccadam{}}: $\Delta_t = (\mathbf{x}^{t} - \textbf{SGD}(\mathbf{x}^{t}, \nabla_{\zeta \sim \mathcal{D}}, \eta_s, N))/\eta_s$
    \State $m_{t+1} \gets (1-\beta_1)\Delta_t + \beta_1m_{t}$
    \State $v_{t+1} \gets (1-\beta_2)\Delta_t^2 + \beta_2v_{t}$
    \State $\hat{m}_{t+1} \gets m_{t+1}/(1-\beta_1^{t+1})$
    \State $\hat{v}_{t+1} \gets v_{t+1}/(1-\beta_2^{t+1})$
    \State $\mathbf{x}^{t+1} \gets \mathbf{x}^{t} - \eta \hat{m}_{t+1}/(\sqrt{\hat{v}_{t+1}}+\epsilon)$
\EndFor
\end{algorithmic}
\end{algorithm}

\begin{algorithm}
\caption{Minibatch Stochastic Gradient Descent (\sgd{})}\label{alg:sgd}
\begin{algorithmic}
\Require model weights $\mathbf{x}^0$, learning rate $\eta_s$, batch size \(\zeta\) and number of steps $N$
\For{step $n$ in $\{0, 1, ... N-1\}$}
    \State Sample a batch of data with size of \(\zeta\) from the training dataset
    \State Calculate gradient estimation \(\nabla_{\zeta \sim \mathcal{D}} F(\mathbf{x}^n) \)
    \State Update model \(\mathbf{x}^{n+1} = \mathbf{x}^n - \eta_s \nabla_{\zeta \sim \mathcal{D}} F(\mathbf{x}^n)\)
\EndFor
\end{algorithmic}
\end{algorithm}

Pseudo codes for \fedopt{} (\cite{reddi2020adaptive}) is given as Algorithm \ref{alg:fedopt}.
\begin{algorithm}[H]
\small
\caption{Adaptive federated optimization (\fedopt{})}\label{alg:fedopt}
\begin{algorithmic}
\Require Seed model \(\mathbf{x}^0\)
    \For{round \(t\) in \(\{0, 1, ... T-1\}\)}
        \For{client \(k\) in \(\{0, 1, ... K-1\}\) \textbf{parallel}} 
        \State \(\mathbf{x}^t_k :=\) \(\textbf{ClientOpt}(\mathbf{x}^t)\) 
        \Comment{Client-side}
        \State \(\Delta^t_k := \mathbf{x}^t_k - \mathbf{x}^t\)
        \EndFor
    \State \(\Delta^t := \textbf{Aggre}(\{\Delta_k^t, 0\leq k < K\})\) \Comment{Server-side}
    \State \(\mathbf{x}^{t+1} := \textbf{ServerOpt}(\Delta^t)\)
    \EndFor
\end{algorithmic}
\end{algorithm}

\newpage
% \appendix
\section{Experiments}
\subsection{Experimental details}\label{subsec:exp_details}
\begin{figure*}[htbp]
\centering
\includegraphics[width=0.2\textwidth]{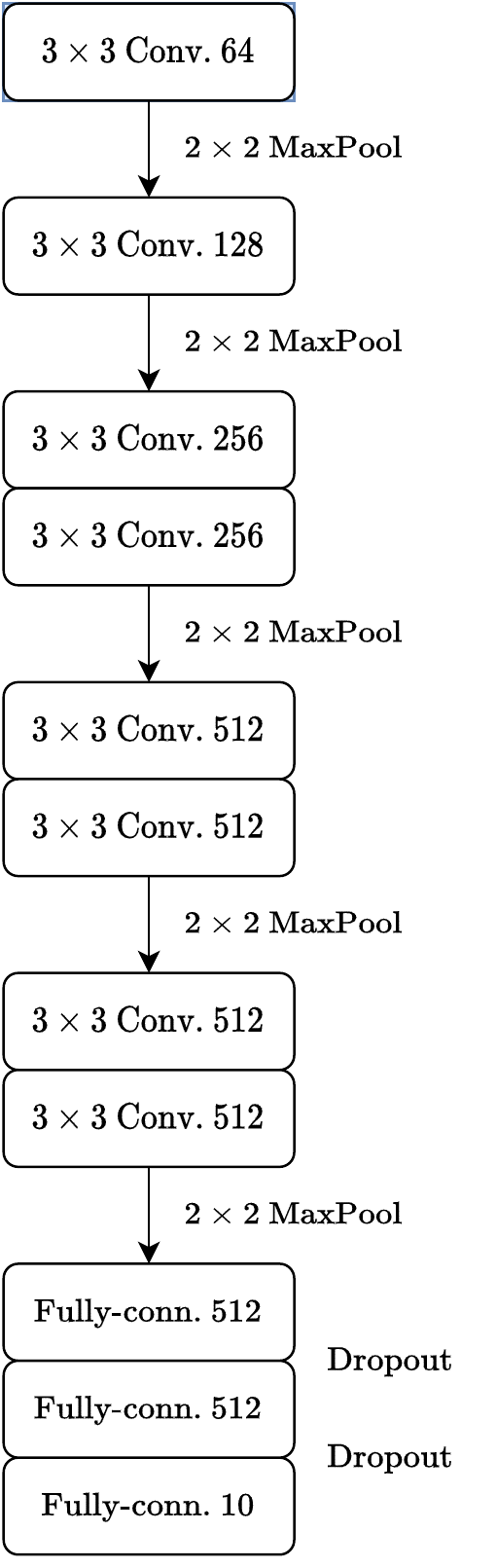}
\caption{Architecture of VGG11 for the CIFAR10 setup}
\label{fig:appen_cifar10_arch}
\end{figure*}

\begin{figure*}[htbp]
\centering
\includegraphics[width=0.12\textwidth]{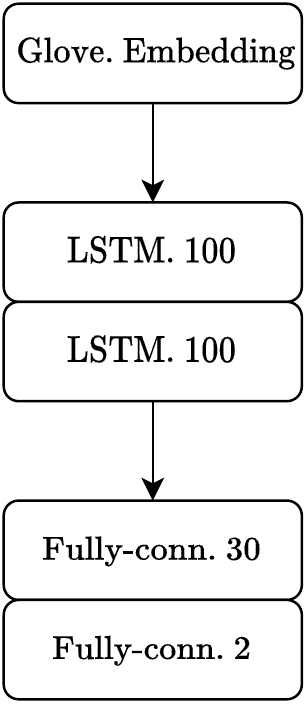}
\caption{Architecture of the two-layer LSTM for the Sent140 setup}
\label{fig:appen_sent_arch}
\end{figure*}

\paragraph{Platform}
All experiments in the paper are conducted on a server with Intel(R) Xeon(R) Gold 6230R CPU and and 2x NVidia RTX A5000 GPUs. All codes are implemented in \textit{PyTorch}.  
\paragraph{Setups}
Details of all federated setups are shown as follows: 
\begin{itemize}
    \item \textbf{Femnist:} A multi-layer perceptron network (MLP) for the classification of the EMNIST dataset. The MLP used for the setup consisted 128 hidden nodes activated by ReLu functions with a loss function of cross-entropy. The EMNIST dataset is partitioned according to the writer of images and each partition acts as a local dataset for each client. Local datasets are thus intrinsically non-IID due to different writing characteristics from different writers. 
    \item \textbf{CIFAR10:} A VGG11 (\cite{simonyan2014very}) model for CIFAR10 dataset. The model used for the setup is VGG11 with slight modifications to be compatible with CIFAR10 dataset. The architecture of the model is shown as Figure \ref{fig:appen_cifar10_arch} with a loss function of cross-entropy. The CIFAR10 dataset is partitioned into 16 subsets by the Dirichlet distribution \(\mathbf{Dir}_{16}(0.05)\) over labels. 
    \item \textbf{Sent140:} An LSTM model (\cite{gers2000learning}) for the sentiment analysis for the Sent140 dataset (\cite{go2009twitter}). Input words are embedded with pretrained Glove (\cite{pennington2014glove}) and logits are output after two LSTM layers with 100 hidden units and one dense layer, with architecture shown in Figure \ref{fig:appen_sent_arch}. The partitioning of the Sent140 dataset follows \cite{caldas2018leaf} and a collection of tweets from each twitter account acts as the local dataset of one client.
    \item \textbf{Synthetic:} A linear regression classifier for multi-class classification on a synthetic dataset, proposed by \cite{caldas2018leaf} as a challenging task for the benchmark of federated learning algorithms. The model is \(y=\textrm{argmax}(\textrm{softmax}(\mathbf{W}x+b))\), where \(x\in \mathbb{R}^{60}\), \(\mathbf{W}\in \mathbb{R}^{10\times60}\) and \(b\in \mathbb{R}^{10}\) with a loss function of cross-entropy. In the Synthetic dataset, there are 100 partitions, the sizes of which follow a power law. 
\end{itemize}

For all setups, each client is associated with a partition and randomly split the local partition with a ratio of \(8:2\) acting as its local training and testing set before federated training starts. 

\paragraph{Hyperparameter settings} For all experiments without specifications, local optimizers for clients are fixed as \sgd{}, with the default Learning rate \(\eta_c=0.01\) for all setups. In each communication round, clients train the model for 1 epoch with batch size 10 in the Femnist, Sent140 and Synthetic setup, and 2 epochs with batch size 32 in the CIFAR10 setup. If local optimizers are set as \sgd{} with (Nesterov) momentum, the momentum factor is fixed as \(0.9\) by default. For server optimizers, \fedavg{} has the default learning rate \(\eta=1\), \fedadam{} has the default hyperparameter set (\(\eta=0.001, \beta_1=0.9\) and \(\beta_2=0.999\)), and \qfedavg{} has learning rate \(\eta=1\) and \(q=1\). Total communication rounds are 500, 200, 1000 and 1000 for the Femnist, Cifar10, Sent140 and Synthetic setup, respectively. For each experiment, global models are initialized with 3 different random seeds and trained independently, and averaged metrics are reported. 

\subsection{Full experimental results}\label{subsec:full_results}
\paragraph{Convergence \& Fairness}
Table \ref{tab:appen_results} shows the full results of the experiment of fairness and convergence. 

\begin{table}[htbp]%{L}{0.6\textwidth}
  \caption{Full experimental results of convergence and fairness: Statistics of test accuracy on clients for \ours{} compared to \fedavg{}, \fedadam{}, \fednova{} and \qfedavg{} with \(q=1\), for Femnist, Cifar10, Sent140 and Synthetic setups.}
  \label{tab:appen_results}
  \centering
  \begin{tabular}{l|l|lll}
    \hline
    Settings & Algorithms & Avg.(\%) & STD.(\%) & Worst 30\%(\%) \\
    \hline
    Femnist &\fedavg & 77.77 \textpm 0.64 & 13.20 \textpm 0.91 & 60.11 \textpm 2.19\\
            &\fedadam & 82.97 \textpm 0.26 & 11.44 \textpm 0.76 & 67.65 \textpm 1.80\\
            &\qfedavg & 76.91 \textpm 0.21 & 10.94 \textpm 0.26 & 64.06 \textpm 0.37\\
            &\fednova & 78.31 \textpm 0.53 & 10.77 \textpm 0.36 & 64.97 \textpm 1.01\\
            &\ours & 84.48 \textpm 0.50 & 8.62 \textpm 0.25 & 74.16 \textpm 0.30\\
    \hline
    Cifar10 &\fedavg & 36.47 \textpm 0.75 & 20.28 \textpm 0.90 & 9.45 \textpm 3.51\\
            &\fedadam & 56.33 \textpm 0.96 & 11.77 \textpm 1.99 & 40.4 \textpm 5.13\\
            &\qfedavg & 28.01 \textpm 0.81 & 21.92 \textpm 0.47 & 3.15 \textpm 3.25\\
            &\fednova & 36.25 \textpm 0.88 & 24.34 \textpm 1.34 & 5.00 \textpm 3.24\\
            &\ours & 62.81 \textpm 1.02 & 8.18 \textpm 1.33  &46.01 \textpm 2.23\\
    \hline
    Sent140 &\fedavg & 62.71 \textpm 3.20 & 21.97 \textpm 3.09 & 36.21 \textpm 7.34\\
            &\fedadam & 69.25 \textpm 0.26 & 18.95 \textpm 0.31 & 46.48 \textpm 0.55\\
            &\qfedavg & 57.29 \textpm 0.82 & 24.22 \textpm 2.50 & 26.89 \textpm 2.83\\
            &\fednova & 63.20 \textpm 3.03 & 22.48 \textpm 2.09 & 35.61 \textpm 6.31\\
            &\ours & 68.90 \textpm 0.31 & 18.63 \textpm 0.52 & 46.82 \textpm 1.13\\
    \hline
    Synthetic   &\fedavg & 88.34 \textpm 0.55 & 16.77 \textpm 0.44 & 25.94 \textpm 1.30\\
                &\fedadam & 89.71 \textpm 0.47 & 14.57 \textpm 0.78 & 57.15 \textpm 10.56\\
                &\qfedavg & 90.04 \textpm 0.66 & 12.48 \textpm 0.76 & 76.50 \textpm 1.50\\
                &\fednova & 92.20 \textpm 0.16 & 10.96 \textpm 0.09 & 83.41 \textpm 1.47\\
                &\ours    & 94.18 \textpm 0.45 & 8.52 \textpm 0.37 & 87.07 \textpm 2.28\\
    \hline
  \end{tabular}
\end{table}

\begin{figure*}[ht]
\centering
\includegraphics[width=0.45\textwidth]{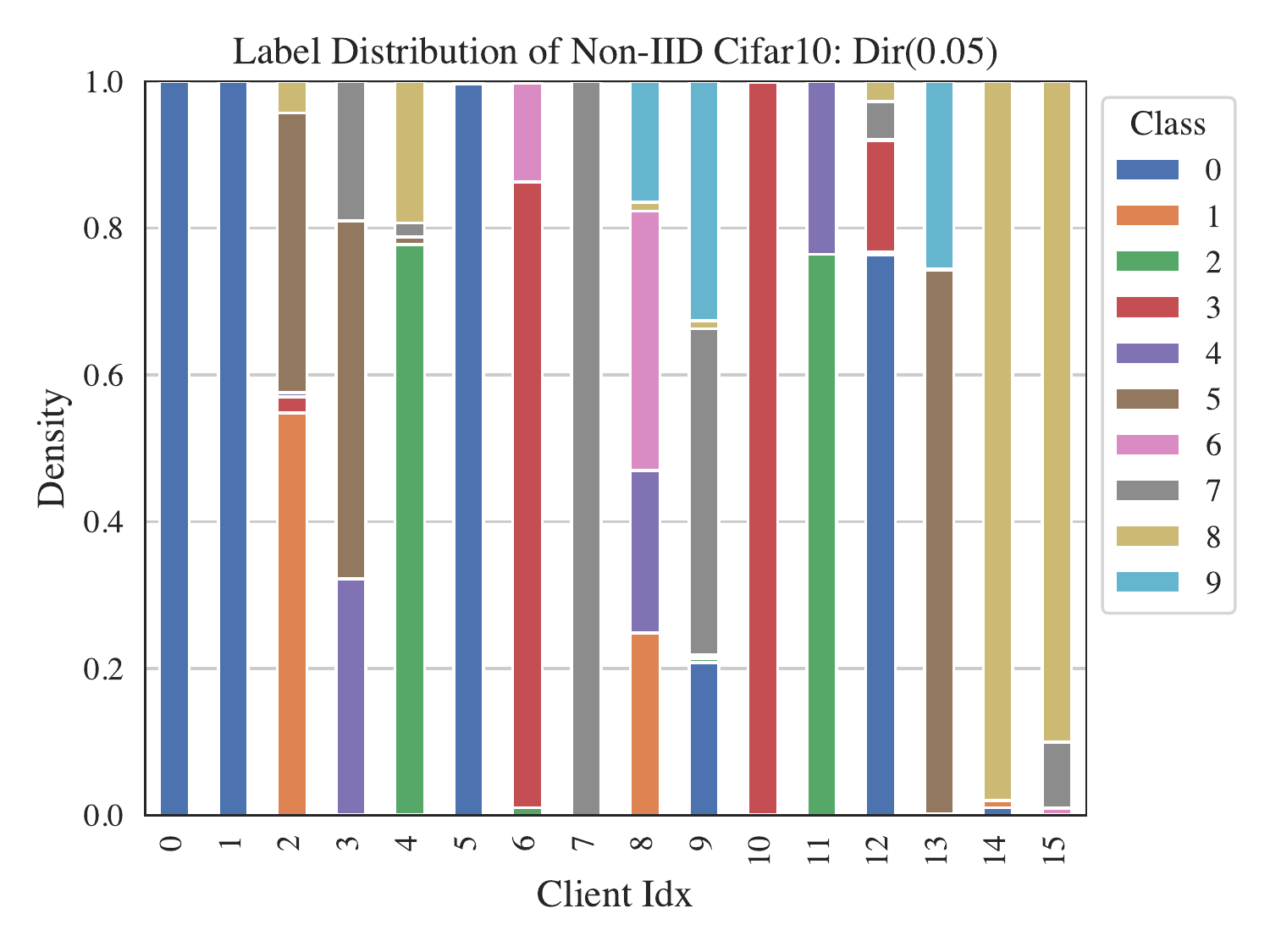}
\includegraphics[width=0.45\textwidth]{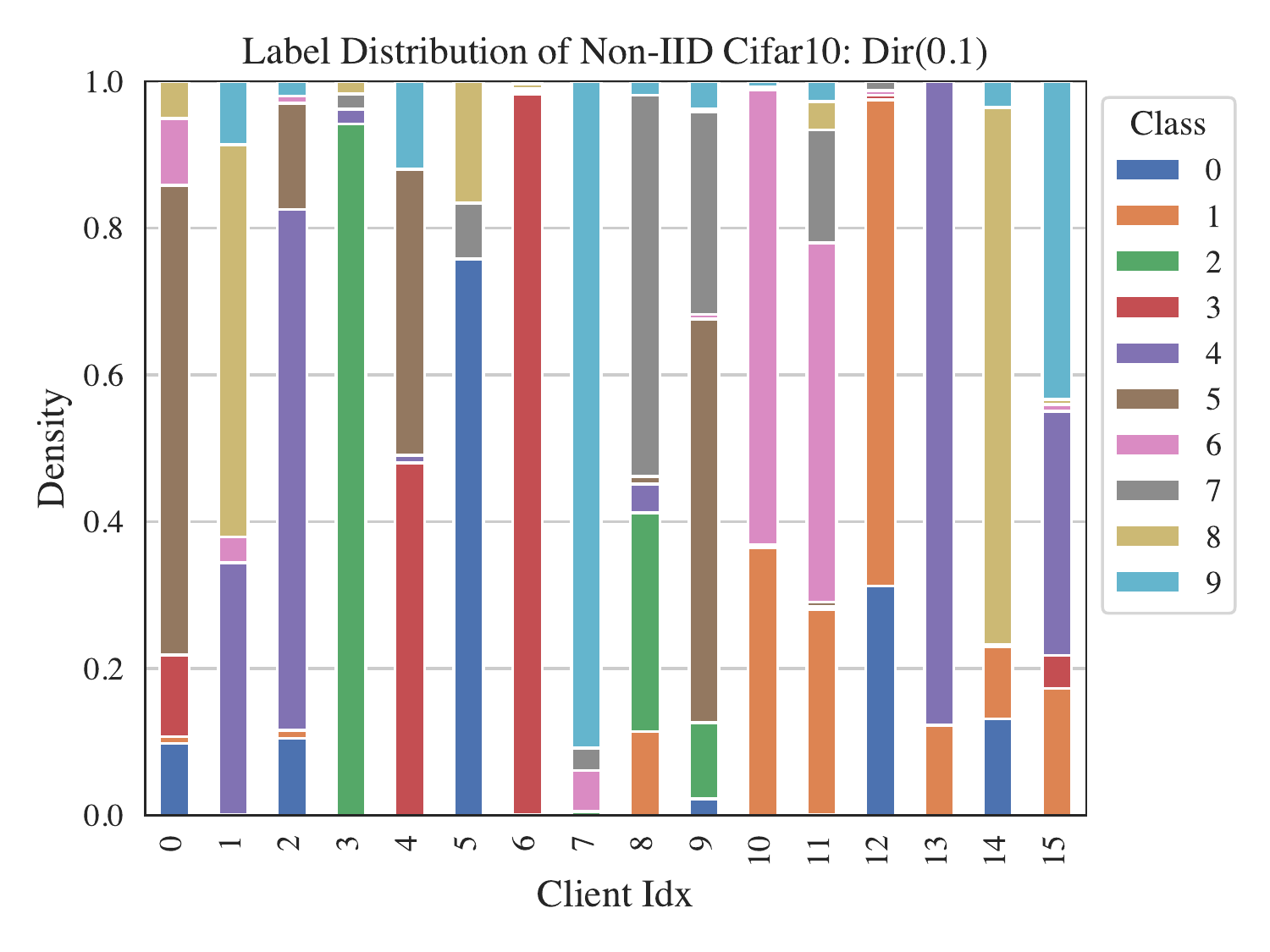}
\includegraphics[width=0.45\textwidth]{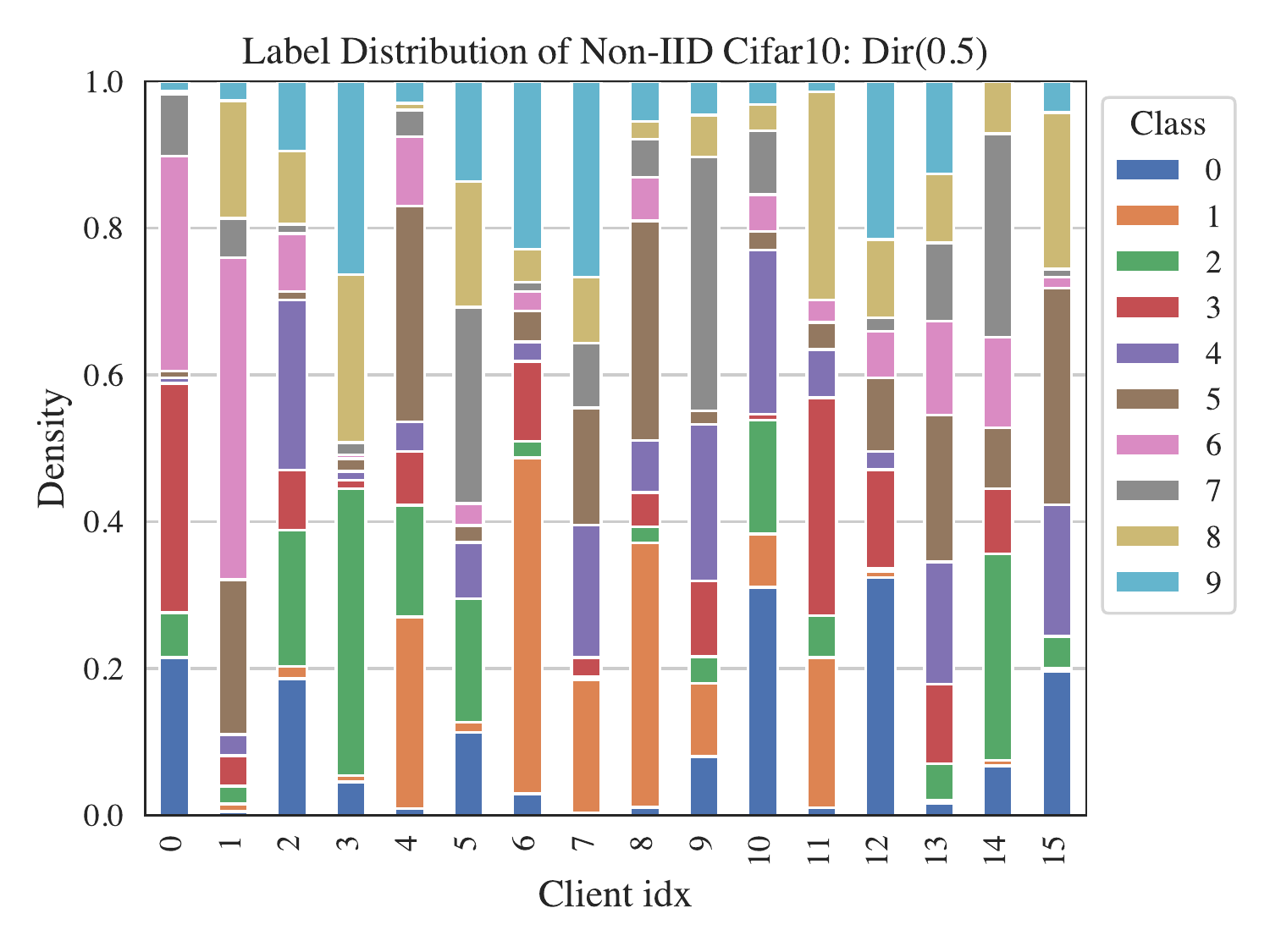}
\includegraphics[width=0.45\textwidth]{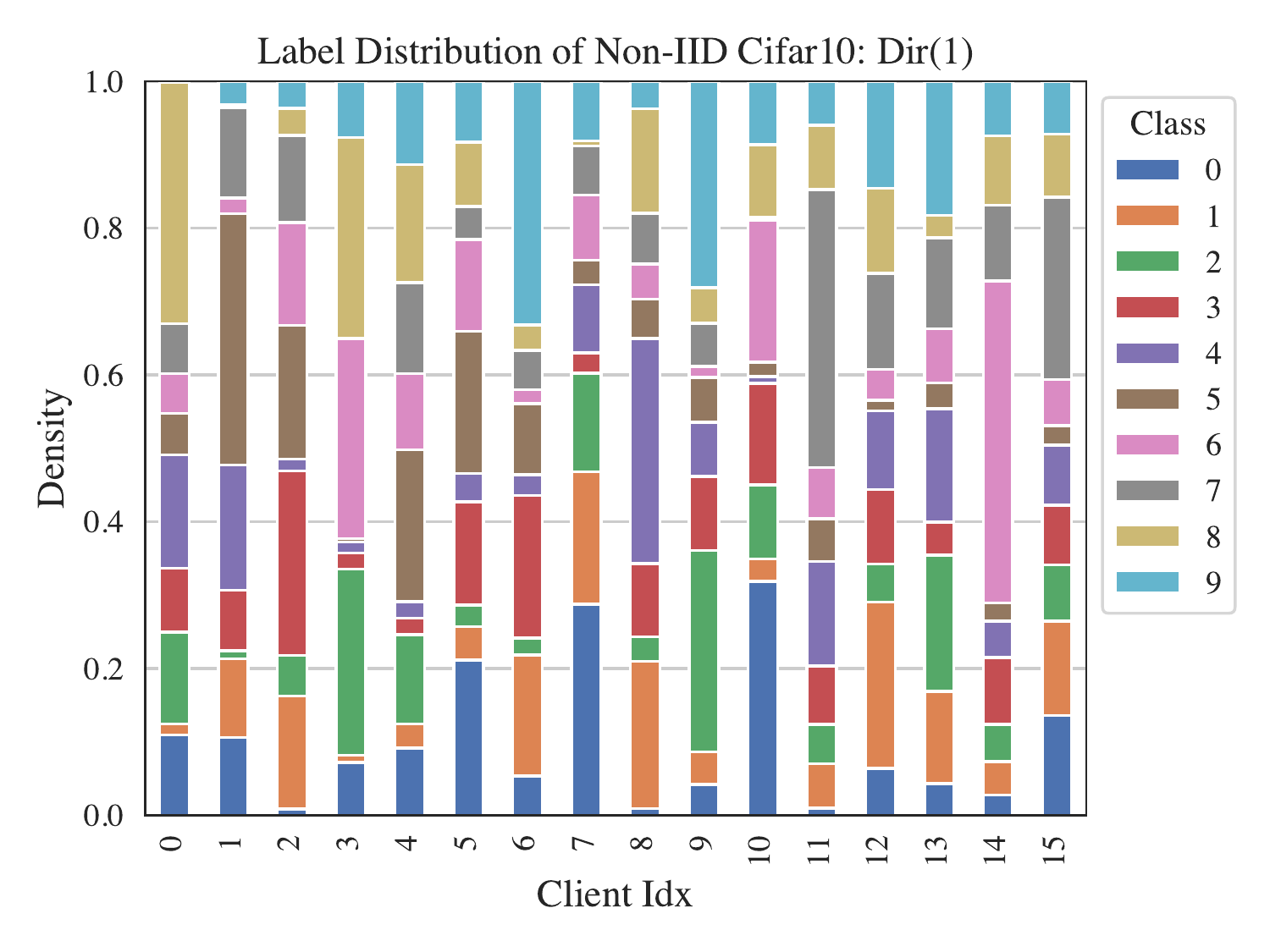}

\caption{Label distributions of local datasets of the Cifar10 setup for different non-IID levels.}
\label{fig:appen_data_dist}
\end{figure*}

\paragraph{Robustness against different levels of data heterogeneity}
Figure \ref{fig:appen_data_dist} shows label distributions of different non-IID levels of the Cifar10 setup. Table \ref{tab:appen_rob_noniid} shows the full results of comparison between different algorithms on the Cifar10 setup. Different levels of data heterogeneity are generated with Dirichlet distribution of different concentration parameter \(\beta\) ranging from \(0.05\) to \(0.5\) and together with an IID partitioning. It can be observed that \ours{} consistently outperforms other algorithms with the highest test accuracy and lowest STD of test accuracy in all different settings. 

\begin{table}[ht]
% \scriptsize
  \caption{Full experimental results of federated algorithms against different levels of data heterogeneity on the Cifar10 setup. }
  \label{tab:appen_rob_noniid}
  \centering
  \begin{tabular}{l|l|l|lll}
    \hline
    Data Distribution & Algorithm & Avg.(\%) & STD.(\%) & Worst 30\%(\%) \\
    \hline
    \textbf{Dir}(0.05)  &\fedavg & 36.47 \textpm 0.75 & 20.28 \textpm 0.90 & 9.45 \textpm 3.51\\
                        &\fedadam & 56.33 \textpm 0.96 & 11.77 \textpm 1.99 & 40.4 \textpm 5.13\\
                        &\qfedavg & 28.01 \textpm 0.81 & 21.92 \textpm 0.47 & 3.15 \textpm 3.25\\
                        &\fednova & 36.25 \textpm 0.88 & 24.34 \textpm 1.34 & 5.00 \textpm 3.24\\
                        &\ours & 62.81 \textpm 1.02 & 8.18 \textpm 1.33  &46.01 \textpm 2.23\\
    \hline
    \textbf{Dir}(0.1)   &\fedavg & 50.41 \textpm 0.46 & 13.21 \textpm 0.36 & 33.20 \textpm 3.98\\
                        &\fedadam & 65.79 \textpm 0.91 & 8.61 \textpm 0.51 & 55.92 \textpm 2.36\\
                        &\qfedavg & 38.95 \textpm 0.73 & 12.46 \textpm 0.20 & 24.59 \textpm 2.11\\
                        &\fednova & 48.09 \textpm 1.82 & 14.29 \textpm 0.58 & 33.34 \textpm 3.28\\
                        &\ours & 66.16 \textpm 1.13 & 8.59 \textpm 0.39 & 56.48 \textpm 1.45\\
    \hline
    \textbf{Dir}(0.5)   &\fedavg & 49.38 \textpm 0.92 & 7.29 \textpm 2.60 & 41.22 \textpm 3.25\\
                        &\fedadam & 70.49 \textpm 0.78 & 3.97 \textpm 0.49 & 65.83 \textpm 0.84\\
                        &\qfedavg & 44.95 \textpm 0.41 & 4.60 \textpm 0.51 & 39.73 \textpm 0.61\\
                        &\fednova & 49.47 \textpm 1.06 & 6.09 \textpm 2.19 & 43.00 \textpm 3.26\\
                        &\ours & 71.43 \textpm 0.81 & 5.40 \textpm 0.22 & 64.93 \textpm 1.04\\
    \hline
    \textbf{Dir}(1):    &\fedavg & 40.97 \textpm 0.66 & 4.93 \textpm 0.47 & 35.53 \textpm 1.12\\
                        &\fedadam & 71.22 \textpm 0.17 & 2.95 \textpm 0.27 & 68.01 \textpm 0.02\\
                        &\qfedavg & 36.27 \textpm 0.95 & 5.40 \textpm 0.75 & 30.63 \textpm 1.56\\
                        &\fednova & 40.28 \textpm 0.10 & 4.70 \textpm 0.49 & 35.30 \textpm 0.40\\
                        &\ours & 72.77 \textpm 0.44 & 3.05 \textpm 0.11 & 69.57 \textpm 0.54\\
    \hline
  \end{tabular}
\end{table}
\paragraph{Compatibility with local momentum}
Table \ref{tab:appen_rob_localmom} shows the full results of different federated algorithms with different local solvers. It is observed that \ours{} is not only compatible with momentum-based local solvers, it also provides better results compared to other federated algorithms.
\begin{table}[ht]
% \scriptsize
  \caption{Full experimental results of federated algorithms in cooperation with local momentum on the Synthetic setup}
  \label{tab:appen_rob_localmom}
  \centering
  \begin{tabular}{l|l|l|lll}
    \hline
    Local Solver & Algorithm & Avg.(\%) & STD.(\%) & Worst 30\%(\%) \\
    \hline
    Vanilla \sgd{}&\fedavg & 88.34 \textpm 0.55 & 16.77 \textpm 0.44 & 25.94 \textpm 1.30\\
                &\fedadam & 89.71 \textpm 0.47 & 14.57 \textpm 0.78 & 57.15 \textpm 10.56\\
                &\qfedavg & 90.04 \textpm 0.66 & 12.48 \textpm 0.76 & 76.50 \textpm 1.50\\
                &\fednova & 92.20 \textpm 0.16 & 10.96 \textpm 0.09 & 83.41 \textpm 1.47\\
                &\ours    & 94.18 \textpm 0.45 & 8.52 \textpm 0.37 & 87.07 \textpm 2.28\\
    \hline
    \sgd{} with Momen.  &\fedavg & 95.26 \textpm 0.22 & 8.42 \textpm 0.24 & 68.65 \textpm 0.19\\
                        &\fedadam & 91.60 \textpm 0.32 & 12.32 \textpm 0.66 & 59.52 \textpm 4.79\\
                        &\qfedavg & 94.64 \textpm 0.22 & 5.73 \textpm 0.03 & 88.04 \textpm 0.02\\
                        &\fednova & 96.12 \textpm 0.09 & 3.82 \textpm 0.05 & 93.07 \textpm 0.15\\
                        &\ours & 97.19 \textpm 0.11 & 3.32 \textpm 0.02 & 93.41 \textpm 0.11\\
    \hline
    \sgd{} with Neste. Momen. &\fedavg & 95.24 \textpm 0.14 & 8.34 \textpm 0.05 & 68.80 \textpm 0.36\\
                        &\fedadam & 91.79 \textpm 0.19 & 12.07 \textpm 0.43 & 61.51 \textpm 3.52\\
                        &\qfedavg & 94.56 \textpm 0.02 & 5.80 \textpm 0.12 & 87.85 \textpm 0.02\\
                        &\fednova & 96.85 \textpm 0.04 & 3.80 \textpm 0.30 & 94.02 \textpm 0.11\\
                        &\ours & 97.27 \textpm 0.16 & 3.19 \textpm 0.21 & 94.19 \textpm 0.24\\

    \hline
  \end{tabular}
\end{table}

\end{document}